\documentclass{article}

\usepackage[preprint]{neurips_2026}

\usepackage[utf8]{inputenc} 
\usepackage[T1]{fontenc}    
\usepackage{hyperref}       
\usepackage{url}            
\usepackage{booktabs}       
\usepackage{amsfonts}       
\usepackage{nicefrac}       
\usepackage{microtype}      
\usepackage{xcolor}         

\usepackage{enumitem}   
\usepackage{booktabs}
\usepackage{multirow}
\usepackage{graphicx}
\usepackage{tcolorbox}
\usepackage{wrapfig}
\usepackage{algorithm}
\usepackage{algpseudocode}
\usepackage{amsmath}

\usepackage{caption}
\usepackage{placeins}

\newcommand{\ours}{SCTab-Diff}
\title{Beyond Distribution Matching: Semantics-Consistent Tabular Diffusion with Weak Semantic Priors}

\author{%
  \textbf{Yili Wang}$^{1}$\thanks{Equal contribution.}
  \quad
  \textbf{Ruxue Shi}$^{1}$\footnotemark[1]
  \quad
  \textbf{Mengnan Du}$^{2}$ 
   \quad
  \textbf{Hangting Ye}$^{1}$
  \quad
  \textbf{Yi Chang}$^{1}$\thanks{Corresponding authors.}
  \quad
  \textbf{Xin Wang}$^{1}$\footnotemark[2] \\[0.5em]
  $^{1}$School of Artificial Intelligence, Jilin University, Changchun, China \\
  $^{2}$School of Artificial Intelligence, The Chinese University of Hong Kong, Shenzhen, China \\
  \texttt{\{yichang,xinwang\}@jlu.edu.cn}
}

\begin{document}

\maketitle

 \begin{abstract}
Synthetic tabular data can match real data distributions while still violating the semantic constraints that govern valid tabular rows. This reveals a key limitation of existing tabular generators: they mainly optimize distributional fidelity, but do not explicitly model weak semantic priors encoded in tabular schema and textual descriptions. In this paper, we propose \ours, a semantics-consistent tabular diffusion framework for high-fidelity synthetic data generation under weakly specified semantic priors. \ours\ first constructs two types of priors, namely intra-column semantics and inter-column symbolic rules, with LLM-assisted extraction from metadata and validation on the real training split. These priors are then used as generation conditions rather than post-hoc filters. Specifically, \ours\ maps heterogeneous column values, column identities, and semantic priors into a unified semantic space, and performs column-wise forward corruption and prior-conditioned reverse denoising to preserve both marginal distributions and rule-consistent cross-column dependencies. Extensive experiments on six real-world tabular benchmarks show that \ours\ consistently improves distributional fidelity, semantic consistency, and downstream task utility over representative VAE-, GAN-, LLM-, and diffusion-based baselines. Additional analyses further demonstrate the robustness of \ours\ when semantic priors are partially unavailable. Code is \url{https://anonymous.4open.science/status/SCTab-Diff-B6DF}. 
\end{abstract}

\section{Introduction}

\label{sec:intro}
\begin{figure}[t]
\centering
\includegraphics[width=1\textwidth]{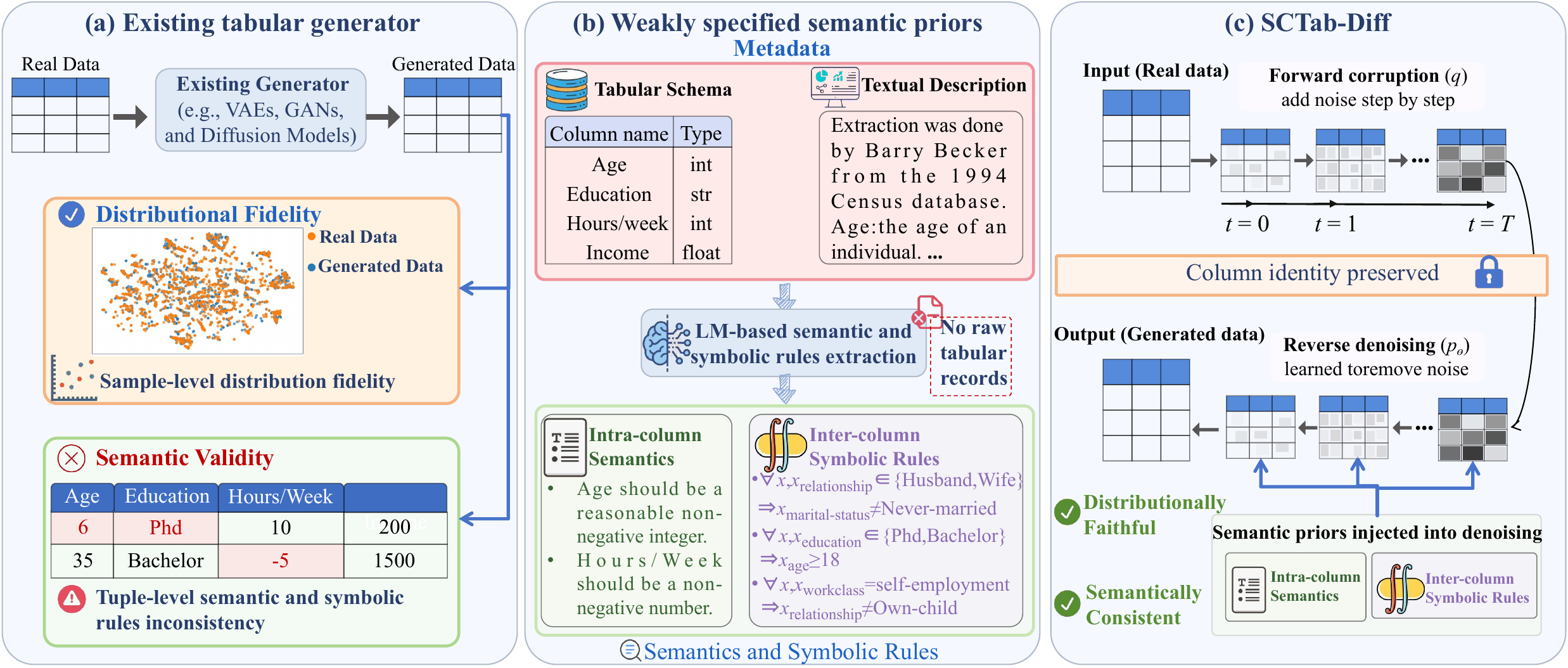}
\vspace{-15pt}
\caption{\label{figure:motivation}From Distributional Fidelity to Semantically Valid Tabular Generation.} 
\vspace{-15pt}
\end{figure}

High-fidelity tabular generation is often reduced to distribution matching~\cite{liu2024entity,schreyer2024imb}, yet for tabular data, \emph{distributional realism alone is not sufficient}~\cite{shi2025tabdiff}.
As illustrated in Fig.~\ref{figure:motivation}(a), a synthetic tuple may appear statistically plausible while still being semantically invalid, for example by assigning an implausible education level to a child or producing values that violate basic cross-column dependency rules. 
Such errors are especially problematic in tabular domains because synthetic data is often reused in downstream pipelines, including data augmentation~\cite{zhangmixed,kotelnikov2023tabddpm}, privacy-preserving data sharing~\cite{wang2024harmonic,torfi2022differentially}, and model training or evaluation~\cite{he2024flexible,sattarov2023findiff,zhang2023adavis}. Therefore, realistic tabular generation should satisfy a stronger objective: generated samples should be both \emph{distributionally faithful} and \emph{semantically valid}.

This requirement is fundamentally tied to the structure of tabular data. Unlike images or text, a table is composed of semantically typed columns rather than interchangeable dimensions~\cite{seedat2023curated}. Each column exhibits its own marginal behavior, while valid tuples must also respect complex dependencies across columns~\cite{kim2024epic}. Moreover, the semantic knowledge needed to characterize valid tables is often only \emph{weakly specified}. As shown in Fig.~\ref{figure:motivation}(b), intra-column semantics are usually implicit in column names, tabular schema, and textual descriptions, whereas inter-column validity is governed by \textbf{symbolic rules} that are rarely explicitly annotated in standard tabular benchmarks. This makes semantically valid tabular generation a substantially harder problem than pure distribution matching. \textbf{This motivates a stronger formulation of tabular generation:} 

\begin{tcolorbox}[
    colback=blue!3,
    colframe=blue!55!black,
    boxrule=0.8pt,
    arc=2mm,
    left=1.5mm,
    right=1.5mm,
    top=1mm,
    bottom=1mm
]
\textbf{Realistic tabular generation requires not only distributional fidelity, but also semantic validity under weakly specified constraints.}

\end{tcolorbox}

Yet this stronger objective is exactly where existing tabular generators fall short. From VAE~\cite{xu2019modeling,ma2020vaem,liu2023goggle} and GAN~\cite{choi2017generating,lee2021invertible,zhao2021ctab} based methods to recent diffusion models~\cite{dhariwal2021diffusion,ho2020denoising,song2020score,austin2021structured}, most existing approaches still follow a \emph{distribution-centric} view of tabular synthesis: each row is treated as a holistic vector, and generation is optimized toward reproducing overall statistics. Such a design is convenient for distribution modeling, but it leaves the generation process largely blind to the semantic structure of tabular data. In particular, existing diffusion formulations corrupt and recover tabular samples without explicitly preserving column identity or conditioning denoising on weak semantic priors. As a result, semantic consistency is treated as an implicit byproduct of distribution learning rather than an explicit generation objective. This limitation is not merely an evaluation issue, but reflects a mismatch between the structure of tabular data and the way it is generated: a tabular row is a set of typed semantic attributes, while most generators model it as an untyped numerical vector.

To bridge this gap, this paper studies \emph{semantically valid tabular generation under weakly specified constraints}. Rather than treating semantic knowledge as external rules for filtering generated samples, we make weak semantic priors part of the diffusion mechanism itself. Specifically, these priors serve as conditioning signals for reverse-time score estimation, so that denoising is guided by both statistical recovery and semantic consistency. As shown in Fig.~\ref{figure:motivation}(c), we construct two types of semantic priors from weak semantic sources: \emph{intra-column semantics}, which encode attribute meaning and value plausibility, and \emph{inter-column symbolic rules}, which capture explicit dependency constraints across attributes. Candidate symbolic constraints are obtained from tabular schema and textual descriptions with the help of LLM-based semantic extraction\footnote{The LLM used for semantic extraction does not access raw tabular records. This design reduces the risk of memorizing or leaking sensitive record-level entries from the original table.}, and are further validated and refined using observed data. Based on this formulation, we propose \textbf{S}emantics-\textbf{C}onsistent \textbf{Tab}ular \textbf{Diff}usion (\ours), which couples column-wise diffusion with prior-conditioned score estimation in a unified semantic space. Within this space, column identity specifies what a noisy latent variable represents, row context specifies whether it is compatible with other attributes, and weak semantic priors specify which recoveries are semantically admissible. This design enables \ours\ to generate synthetic tables that are both distributionally faithful and semantically consistent. 

Our main contributions are summarized as follows:
\begin{itemize}[leftmargin=*, itemsep=2pt, topsep=2pt, parsep=0pt, partopsep=0pt]
    \item We formulate tabular generation beyond distribution matching, requiring synthetic tables to satisfy both \textbf{distributional fidelity} and \textbf{semantic validity} under weakly specified constraints.
    
    \item We introduce a weak semantic prior construction strategy that derives \textbf{intra-column semantics} for attribute-level plausibility and \textbf{inter-column symbolic rules} for cross-attribute dependency validity from tabular schema and textual descriptions.
    
    \item We propose \textbf{\ours}, a semantics-consistent diffusion framework that treats weak semantic priors as generation conditions and injects them into score-based denoising to preserve column identity and enforce rule-consistent dependencies.
    
    \item Extensive experiments demonstrate improvements in distributional fidelity, semantic consistency, and downstream utility, with additional robustness under incomplete semantic priors.
\end{itemize}

\section{Preliminaries}

\subsection{Problem Formulation}
\vspace{-6pt}
Let \(D=\{X,F\}\) denote a tabular dataset, where \(X=\{x^{(m)}\}_{m=1}^{M}\) is the set of table rows and \(F=\{f_i\}_{i=1}^{n}\) is the set of column names. Each row \(x\in X\) is represented as \(x=(x_1,\dots,x_n)\), where the \(i\)-th attribute \(x_i\) corresponds to column \(f_i\). Among the \(n\) columns, \(N_n\) are numerical and \(N_c\) are categorical, with \(N_n+N_c=n\).

In addition to \(D\), this work considers weak semantic sources, such as tabular schema and accompanying textual documentation, from which semantic priors are constructed:
\[
P=\{P^{\mathrm{intra}},P^{\mathrm{inter}}\}.
\]
Here, \(P^{\mathrm{intra}}\) denotes intra-column semantics and \(P^{\mathrm{inter}}\) denotes inter-column symbolic rules. The goal is to learn a generator \(g_\theta(D,P)\) that produces synthetic samples \(\hat{x}\in\hat{D}\) that are both distributionally faithful to \(D\) and semantically valid under \(P\).

\vspace{-10pt}
\subsection{Score-Based Diffusion Preliminaries}
\vspace{-6pt}
\label{sub:sbdm}
Score-based diffusion models perturb data with noise and learn a score function to reverse the corruption process~\cite{song2020score,song2020improved}. This work adopts the variance exploding (VE) formulation~\cite{song2020score}, where the forward perturbation admits the closed form:
\begin{equation}
x_t=x_0+\sigma(t)\epsilon,\qquad \epsilon\sim\mathcal{N}(0,I).
\label{eq:fp}
\end{equation}
The corresponding reverse-time denoising process is:
\begin{equation}
\mathrm{d}x_t=
-2\,\dot{\sigma}(t)\sigma(t)\nabla_{x_t}\log p_t(x_t)\,\mathrm{d}t
+\sqrt{2\,\dot{\sigma}(t)\sigma(t)}\,\mathrm{d}\bar{w}_t.
\label{eq:rp}
\end{equation}
where \(p_t\) denotes the perturbed data distribution at time \(t\), and \(\bar{w}_t\) is a standard reverse-time Wiener process. In this work, the VE formulation serves as the diffusion backbone of \ours, upon which semantics-consistent tabular denoising is built.

\vspace{-5pt}
\section{Proposed method: \ours}
\vspace{-5pt}

\begin{figure*}[t]
\centering
\makebox[\textwidth][c]{%
  \includegraphics[width=1.2\textwidth]{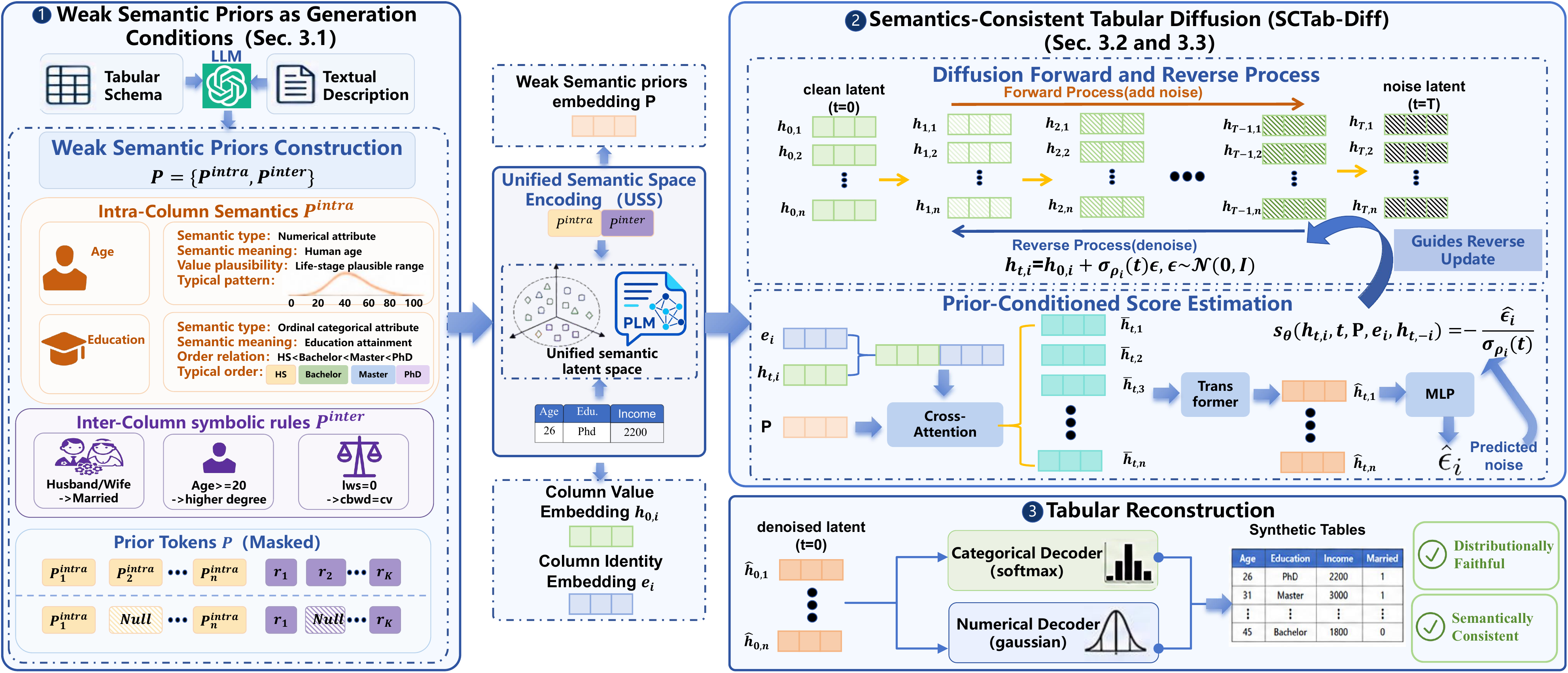}%
}
\vspace{-15pt}
\caption{\label{figure:framework} An overview of \ours.}
\vspace{-15pt}
\end{figure*}

The central idea of \ours\ is to turn weakly specified semantic knowledge into explicit generation conditions for tabular diffusion. Unlike conventional tabular generators that mainly recover the data distribution from noisy samples, \ours\ conditions the reverse denoising process on semantic priors that describe both attribute-level plausibility and cross-attribute dependency rules. As illustrated in Fig.~\ref{figure:framework}, we first formalize these weak semantic priors as generation conditions, then reformulate score-based diffusion to incorporate them into reverse-time denoising, and finally instantiate the formulation in a unified semantic space for end-to-end tabular synthesis.

\vspace{-5pt}
\subsection{Weak Semantic Priors as Generation Conditions}
\vspace{-5pt}
\label{sub:priors}

The reverse diffusion process in \ours\ is guided by weak semantic priors \(P\), which serve as generation conditions beyond distributional recovery. These priors encode two complementary aspects of semantic consistency: attribute-level plausibility and tuple-level dependency validity. Formally, we decompose \(P\) as:
\begin{equation}
P=\{P^{\mathrm{intra}},P^{\mathrm{inter}}\},\qquad
P^{\mathrm{intra}}=\{p_i^{\mathrm{intra}}\}_{i=1}^{n},\qquad
P^{\mathrm{inter}}=\{r_k\}_{k=1}^{K}.
\label{eq:prior_set}
\end{equation}

\noindent\textbf{Intra-column semantics.}
Each \(p_i^{\mathrm{intra}}\) describes the semantic validity of an individual column \(f_i\). It defines the plausible value space of this column, such as numerical ranges, categorical domains, or domain-specific value constraints. These priors operate at the attribute level and prevent the generator from producing values that may be numerically possible but semantically implausible.

\noindent\textbf{Inter-column symbolic rules.}
Each \(r_k\in P^{\mathrm{inter}}\) captures a symbolic dependency among multiple columns:
\begin{equation}
r_k:\quad \phi_k(x_{i_1},\dots,x_{i_s})=1,
\label{eq:rule_form}
\end{equation}
where \(\phi_k(\cdot)\) is a Boolean constraint function over a subset of columns. These priors operate at the tuple level and constrain whether different attributes can co-occur consistently, such as whether an education level is compatible with age or whether income is compatible with working hours.
Given \(P\), a synthetic row \(\hat{x}\) is semantically valid if it satisfies both attribute-level and tuple-level priors:
\begin{equation}
\hat{x}\in \mathcal{V}(P)
\Longleftrightarrow
\hat{x}_i \models p_i^{\mathrm{intra}},\ \forall i,
\quad \text{and} \quad
\phi_k(\hat{x})=1,\ \forall r_k\in P^{\mathrm{inter}},
\label{eq:validity_set}
\end{equation}
where \(\mathcal{V}(P)\) denotes the semantic validity set induced by the weak semantic priors.
In practice, \(P\) is rarely provided as explicit annotations in public tabular benchmarks. We construct candidate priors from tabular schema and textual descriptions with LLM-based semantic extraction, and further validate and refine them using observed data. The resulting \(P\) is then used as the condition for semantics-consistent diffusion.

\subsection{From Distribution-Centric Diffusion to Semantics-Consistent Generation}
\label{sub:reformulation}

Given the weak semantic priors \(P\) defined in Section~\ref{sub:priors}, we next reformulate score-based diffusion from distribution-centric recovery to semantics-consistent generation. Let \(z_{0,i}\) denote the continuous representation of the \(i\)-th column value \(x_i\). In Section~\ref{sub:instantiation}, \(z_{0,i}\) will be instantiated as the unified semantic representation \(h_{0,i}\). We first apply column-wise forward corruption:
\begin{equation}
\label{eq:clfp_new}
z_{t,i}=z_{0,i}+\sigma_{\rho_i}(t)\epsilon,\qquad \epsilon\sim\mathcal{N}(0,I),
\end{equation}
where \(\sigma_{\rho_i}(t)\) is a learnable noise schedule for column \(f_i\). Compared with holistic row-wise corruption, this formulation preserves column-level modeling granularity and allows different columns to follow their own perturbation dynamics.

A direct reverse process would estimate the marginal score \(\nabla_{z_{t,i}}\log p_t(z_{t,i})\), which only encourages denoising toward statistically likely regions. However, semantically valid tabular generation requires a stronger target: each column should be recovered under its column identity, the remaining attributes, and the weak semantic priors. We replace the marginal score with a prior-conditioned score:
\begin{equation}
\label{eq:sf_new}
\nabla_{z_{t,i}}\log p_t
\bigl(
z_{t,i}\mid P, f_i, z_{t,-i}
\bigr),
\end{equation}
where \(z_{t,-i}=\{z_{t,j}\}_{j\neq i}\) denotes the noisy representations of other columns. This changes the denoising target from recovering a distributionally plausible column value to recovering a column value that is also compatible with semantic priors and cross-column context.

Since weak semantic priors may be incomplete or partially unavailable, we randomly mask prior tokens during training:
\begin{equation}
\label{eq:m_new}
M=\mathbf{1}\{U>p_u\},\qquad U\sim \mathrm{Uniform}(0,1)^m,
\end{equation}
where \(m\) is the number of prior tokens and \(p_u\) controls the masking probability. The reverse-time process is then written as:
\begin{equation}
\begin{split}
\label{eq:re_new}
\mathrm{d}z_{t,i}={}&
-2\dot{\sigma}_{\rho_i}(t)\sigma_{\rho_i}(t)
\nabla_{z_{t,i}}\log p_t
\bigl(
z_{t,i}\mid P\!\ast\!M, f_i, z_{t,-i}
\bigr)\mathrm{d}t \\
&+\sqrt{2\dot{\sigma}_{\rho_i}(t)\sigma_{\rho_i}(t)}\,\mathrm{d}\omega_t .
\end{split}
\end{equation}

This reformulation incorporates weak semantic priors into the prior-conditioned reverse diffusion process, rather than using them as external post-hoc filters. As a result, the denoising process is guided by both statistical recovery and semantic consistency, enabling generation to preserve column-wise distributions while enforcing rule-consistent inter-column dependencies. Additional discussion on this prior-conditioned denoising mechanism is provided in Appendix~\ref{app:theory_analysis}.

\subsection{Instantiating SCTab-Diff}

\label{sub:instantiation}

The reformulation in Section~\ref{sub:reformulation} defines a prior-conditioned diffusion process over continuous column representations, where denoising is guided by three types of conditions: the column identity \(f_i\), the weak semantic priors \(P\), and the remaining columns \(z_{t,-i}\). This section describes how \ours\ instantiates these abstract conditions in a unified semantic latent space. Specifically, the continuous representation \(z_{0,i}\) is implemented as a semantic embedding \(h_{0,i}\), the column identity \(f_i\) is represented by an embedding \(e_i\), and the semantic prior set \(P\) is encoded as \(\mathbf P\). The denoising network \(s_\theta\) then implements the prior-conditioned score estimation in Eq.~\eqref{eq:sf_new}.

\noindent\textbf{Mapping tabular variables and priors into a unified semantic space(USS).}
Since tabular columns are heterogeneous in type and semantics, directly applying diffusion to raw values is not suitable, especially for categorical attributes. Therefore, \ours\ first maps each column value and its column identity into a shared semantic space. For the \(i\)-th column, the initial latent representation is defined as:
\begin{equation}
h_{0,i} =
\begin{cases}
\mathrm{Average}(\mathrm{PLM}([f_i, x_i])), & \text{for categorical columns},\\
\mathrm{Average}(\mathrm{PLM}([f_i])) \cdot x_i, & \text{for numerical columns}.
\end{cases}
\label{eq:semantic_encoding}
\end{equation}
This construction makes \(h_{0,i}\) the concrete instantiation of the continuous variable \(z_{0,i}\) in Section~\ref{sub:reformulation}. For categorical columns, the value is encoded together with its column name to preserve semantic meaning. For numerical columns, the column-name embedding is modulated by the numerical value, so that the representation remains value-sensitive while retaining column semantics.

The other two conditioning signals are instantiated similarly. The column identity \(f_i\) is encoded as:                
\begin{equation}
e_i=\mathrm{Average}(\mathrm{PLM}(f_i)),
\label{eq:column_identity}
\end{equation}
and the weak semantic priors \(P\) defined in Section~\ref{sub:priors} are encoded as
$\mathbf P=\mathrm{PLM}(P)$.
Thus, column values, column identities, and weak semantic priors are all represented in the same latent space. This shared space is what allows \(P\) to act as a generation condition rather than a detached external rule set.

\noindent\textbf{Implementing prior-conditioned denoising.}
Within this latent space, \ours\ applies the column-wise forward process(Cw-FP) to each semantic representation:
\begin{equation}
h_{t,i}=h_{0,i}+\sigma_{\rho_i}(t)\epsilon,\qquad \epsilon\sim\mathcal{N}(0,I).
\label{eq:latent_forward}
\end{equation}
The main role of the denoising network is to approximate the prior-conditioned score:
\begin{equation}
s_\theta(h_{t,i},t,\mathbf P,e_i,h_{t,-i})
\approx
-\sigma_{\rho_i}(t)
\nabla_{h_{t,i}}\log p_t(h_{t,i}\mid \mathbf P,e_i,h_{t,-i}).
\label{eq:score_approx}
\end{equation}
where \(h_{t,-i}=\{h_{t,j}\}_{j\neq i}\) denotes the noisy representations of all columns except the \(i\)-th column. This equation shows how the abstract score in Section~\ref{sub:reformulation} is implemented: \(e_i\) provides column identity, \(\mathbf P\) provides semantic priors, and \(h_{t,-i}\) provides cross-column context.

To preserve column identity under corruption, the noisy representation is first annotated with its column identity:
\begin{equation}
\tilde h_{t,i}=[h_{t,i}\|e_i].
\label{eq:column_annotation}
\end{equation}
This step implements the conditioning on \(f_i\). It prevents the denoising network from treating different columns as interchangeable latent dimensions.

Next, to make weak semantic priors actively guide denoising, \(\mathbf P\) is used as semantic context through cross-attention:
\begin{equation}
\bar h_{t,i}=
 \tilde{h}_{t,i}+\mathrm{Softmax}\!\left(
\frac{(\tilde h_{t,i}W^Q)((\mathbf P\!\ast\!M)W^K)^\top}{\sqrt d}
\right)(\mathbf P\!\ast\!M)W^V,
\label{eq:prior_injection}
\end{equation}
where \(M\) is the prior mask defined in Eq.~\eqref{eq:m_new}. This operation implements the conditioning on \(P\). Instead of using semantic priors as post-hoc filters, \ours\ lets each noisy column representation retrieve relevant prior information during reverse denoising.

Finally, to implement the dependency on other columns \(h_{t,-i}\), the prior-enhanced column representations are passed into a Transformer-based dependency module:
\begin{equation}
\hat h_{t,1},\dots,\hat h_{t,n}
=
\mathrm{DepCapture}(\bar h_{t,1},\dots,\bar h_{t,n}).
\label{eq:dep_capture}
\end{equation}
The resulting representation \(\hat h_{t,i}\) integrates the three conditioning signals in Eq.~\eqref{eq:sf_new}: column identity, weak semantic priors, and cross-column context. We then predict the injected noise by $\hat\epsilon_i=\mathrm{MLP}(\hat h_{t,i})$, and use it to parameterize the prior-conditioned score:
\begin{equation}
s_\theta(h_{t,i},t,\mathbf P,e_i,h_{t,-i})
=
-\frac{\hat\epsilon_i}{\sigma_{\rho_i}(t)}.
\label{eq:score_from_noise}
\end{equation}
This score determines the reverse denoising direction. Since it is conditioned on \(\mathbf P\), \(e_i\), and \(h_{t,-i}\), each denoising step is guided not only by statistical recovery, but also by semantic priors and cross-column context. The denoising network is trained by minimizing the noise prediction error:
\begin{equation}
\mathcal L_{\mathrm{diff}}
=
\mathbb E_{t\sim p(t)}
\sum_{i=1}^{n}
\left\|
\hat\epsilon_i-\epsilon_i
\right\|_2^2.
\label{eq:diff_loss}
\end{equation}

\noindent\textbf{Reconstructing tabular rows.}
After reverse denoising, \ours\ obtains the recovered latent representation \(\hat h_{0,i}\) for each column. The tabular decoder maps it back to the original data space:
\begin{equation}
\hat{x}_i =
\begin{cases}
\mathrm{Softmax}(\hat h_{0,i}w_i^{\mathrm{cat}}+b_i^{\mathrm{cat}}), & \text{for categorical columns},\\
\hat h_{0,i}w_i^{\mathrm{num}}+b_i^{\mathrm{num}}, & \text{for numerical columns},
\end{cases}
\label{eq:decoder}
\end{equation}
where \(w_{i}^{cat} \in \mathbb{R}^{d\text{×}C_i}\), \(b_{i}^{cat} \in \mathbb{R}^{1\text{×}1}\), \(w_{i}^{num} \in \mathbb{R}^{d\text{×}1}\), \(b_{i}^{num} \in \mathbb{R}^{1\text{×}1}\) are decoder’s parameters for categorical and numerical columns.  The decoder is trained with a reconstruction objective:

\begin{equation}
\mathcal L_{\mathrm{rec}}
=
\ell_{\mathrm{recon}}(x,\hat{x}) =
\sum_{i\in\mathcal{C}}
\mathrm{CE}(x_i,\hat{x}_i)
+
\sum_{i\in n}
\|x_i-\hat{x}_i\|_2^2.,
\label{eq:rec_loss}
\end{equation}
where \(\ell_{\mathrm{recon}}\) measures the reconstruction error between the input and reconstructed data. Through this instantiation, the three conditions introduced in the reformulated score function are explicitly realized: \(e_i\) preserves column identity, \(\mathbf P\) injects weak semantic priors, and \(h_{t,-i}\) enables cross-column dependency modeling.  For clarity, Algorithm~\ref{alg:train} presents the training procedure of \ours, while Algorithm~\ref{alg:sample} describes synthetic tabular data generation.

\vspace{-5pt}
\section{Experiments}

\vspace{-5pt}
To comprehensively evaluate the performance of \ours\ we focus on the following key research questions:
\textbf{Q1:} Can \ours\ preserve the distributional fidelity of real tabular data?
\textbf{Q2:} Can \ours\ improve semantic validity by satisfying intra-column semantics and inter-column symbolic rules?
\textbf{Q3:} Can synthetic data generated by \ours\ support downstream predictive tasks?
\textbf{Q4:} How robust is \ours\ when weak semantic priors are incomplete or partially unavailable?

\subsection{Experimental Setups}

\textbf{Datasets.}~We evaluate \ours\ using six real-world datasets from~\cite{zhangmixed}, including four classification datasets (i.e., Adult, Default, Shoppers, and Magic) and two regression datasets (i.e., Beijing and News). 
Refer to Appendix~\ref{suba:datasets} for details of the dataset. \\
\textbf{Baselines.}~We compare \ours\ against seven baseline models that are categorized into three groups: 1) Traditional methods: CTGAN~\cite{xu2019modeling}, CTGAN+~\cite{zhao2024ctab}, and TVAE~\cite{xu2019modeling}; 2) LLM-based methods: P-TA~\cite{yang2024p}; 3) Diffusion-based methods: TabDDPM~\cite{kotelnikov2023tabddpm}, TABSYN~\cite{zhangmixed}, and TABDIFF~\cite{shi2025tabdiff}. Constrained variants based on post-hoc filtering or rule-based repair are discussed in Appendix~\ref{app:constrained_baselines}, since they impose rules after generation and are not directly comparable to our prior-conditioned denoising mechanism.\\
\textbf{Implementation Details.}~We compare \ours\ to all baseline methods using the same experimental setup. All the methods are optimized with the Adam optimizer, and all the experiments are conducted on an Nvidia L40 GPU (48GB) with the same seed set. Our method uses a pretrained BERT-base-uncased~\cite{devlin2019bert} (PLM($\cdot$)) to encode tabular data into a unified semantic space \(\mathbb{R}^d\), where \(d=128\). The batch size is set to 4096, and the learning rate is set to 1e-3. All reported results are averaged over 5 runs with different random seeds, and the subscripts denote standard deviations. \\
\textbf{Evaluation Metrics.}~We evaluate the quality of the generated data from three aspects: \emph{distribution fidelity}, \emph{structural and semantic fidelity}, and \emph{task fidelity}. Details of all metrics are provided in Appendix~\ref{suba:metrics}. 

\vspace{-10pt}
\subsection{Main Results}

\subsubsection{Distributional and Semantic Fidelity}
\label{sub:fidelity}

To answer Q1 and Q2, we evaluate generated data using Shape, Trend, and SA in Table~\ref{tab:data_fidelity}. Shape measures marginal distribution fidelity at the column level. Trend reflects cross-column dependency preservation. SA evaluates whether generated rows satisfy weak semantic priors, including intra-column semantics and inter-column symbolic rules.

\renewcommand{\arraystretch}{0.78}
\begin{table*}[t]
  \centering
  \caption{Data fidelity evaluation results. Shape(\%) $\uparrow$  evaluates marginal distribution, while Trend(\%) $\uparrow$  and SA(\%) $\uparrow$  evaluate data fidelity. $\uparrow$ indicates that higher values are better. The best results are highlighted in bold, and suboptimal results are marked with an underline.}
  \setlength{\tabcolsep}{2.5pt}
  \resizebox{1.0\textwidth}{!}{
    \begin{tabular}{c|c|ccccccc|c}
    \toprule
    Data & Metric & CTGAN & CTGAN+ & TVAE & P-TA & TabDDPM & TABSYN & TABDIFF & \textbf{\ours} \\
    \midrule

    \multirow{3}{*}{Beijing}
    & Shape & 80.59\(_{1.18}\) & 89.30\(_{6.15}\) & 79.32\(_{1.40}\) & 88.07\(_{0.27}\) & 64.01\(_{30.05}\) & 85.24\(_{0.06}\) & \underline{98.35\(_{0.28}\)} & \textbf{98.56\(_{0.20}\)} \\
    & Trend & 80.38\(_{0.62}\) & 77.96\(_{5.60}\) & 81.67\(_{1.68}\) & 76.44\(_{2.89}\) & 58.56\(_{31.97}\) & 76.32\(_{0.03}\) & \underline{96.26\(_{0.23}\)} & \textbf{97.09\(_{0.05}\)} \\
    & SA & 24.75\(_{18.48}\) & 87.84\(_{6.48}\) & 19.92\(_{3.38}\) & 39.02\(_{2.52}\) & 97.03\(_{26.15}\) & 97.58\(_{0.07}\) & \underline{98.60\(_{0.30}\)} & \textbf{98.68\(_{0.05}\)} \\
    \midrule

    \multirow{3}{*}{News}
    & Shape & 83.86\(_{0.66}\) & 49.56\(_{1.74}\) & 83.07\(_{0.39}\) & 96.13\(_{0.03}\) & 13.52\(_{6.21}\) & 96.45\(_{0.07}\) & \underline{96.62\(_{0.29}\)} & \textbf{96.90\(_{0.30}\)} \\
    & Trend & 94.82\(_{0.09}\) & 89.04\(_{0.43}\) & 93.41\(_{0.03}\) & 88.49\(_{0.10}\) & 28.45\(_{47.59}\) & \underline{98.35\(_{0.03}\)} & 98.14\(_{0.32}\) & \textbf{98.60\(_{0.20}\)} \\
    & SA & 0.00\(_{0.00}\) & 0.00\(_{0.00}\) & 0.00\(_{0.00}\) & 0.00\(_{0.00}\) & 0.00\(_{0.00}\) & 0.76\(_{0.04}\) & \textbf{1.42\(_{1.17}\)} & \underline{1.24\(_{0.02}\)} \\
    \midrule

    \multirow{3}{*}{Shoppers}
    & Shape & 77.42\(_{1.78}\) & 70.00\(_{5.41}\) & 76.59\(_{1.23}\) & 83.35\(_{0.01}\) & 97.11\(_{0.64}\) & 94.90\(_{0.03}\) & \textbf{97.94\(_{0.21}\)} & \underline{97.89\(_{0.30}\)} \\
    & Trend & 86.78\(_{0.21}\) & 70.52\(_{3.27}\) & 81.88\(_{1.44}\) & 52.75\(_{0.10}\) & 93.87\(_{1.67}\) & 92.40\(_{0.30}\) & \underline{97.88\(_{0.15}\)} & \textbf{98.00\(_{0.40}\)} \\
    & SA & 5.28\(_{2.29}\) & 94.01\(_{3.93}\) & 10.80\(_{2.05}\) & 0.00\(_{0.00}\) & 97.56\(_{1.49}\) & 97.97\(_{0.10}\) & \underline{99.19\(_{0.09}\)} & \textbf{99.98\(_{0.01}\)} \\
    \midrule

    \multirow{3}{*}{Adult}
    & Shape & 83.78\(_{1.89}\) & 83.34\(_{5.94}\) & 85.10\(_{0.91}\) & 91.20\(_{0.18}\) & \underline{98.94\(_{0.13}\)} & 95.12\(_{0.02}\) & 98.84\(_{0.38}\) & \textbf{99.06\(_{0.12}\)} \\
    & Trend & 83.22\(_{2.54}\) & 82.11\(_{8.21}\) & 85.68\(_{1.41}\) & 76.81\(_{3.07}\) & \underline{97.75\(_{0.52}\)} & 86.73\(_{0.65}\) & 97.20\(_{0.56}\) & \textbf{97.98\(_{0.50}\)} \\
    & SA & 0.00\(_{0.00}\) & 81.59\(_{0.91}\) & 0.00\(_{0.00}\) & 10.16\(_{17.59}\) & \underline{92.66\(_{0.17}\)} & 61.95\(_{0.27}\) & 91.59\(_{0.29}\) & \textbf{93.76\(_{0.60}\)} \\
    \midrule

    \multirow{3}{*}{Default}
    & Shape & 86.14\(_{0.39}\) & 85.56\(_{0.66}\) & 88.92\(_{0.33}\) & 93.36\(_{0.04}\) & 98.35\(_{0.10}\) & 96.57\(_{0.09}\) & \underline{98.37\(_{0.37}\)} & \textbf{98.57\(_{0.20}\)} \\
    & Trend & 76.70\(_{0.83}\) & 82.42\(_{1.55}\) & 80.08\(_{2.75}\) & 27.92\(_{0.93}\) & 94.13\(_{0.26}\) & 87.56\(_{1.55}\) & \underline{96.46\(_{1.26}\)} & \textbf{97.89\(_{0.30}\)} \\
    & SA & 52.56\(_{2.22}\) & 98.78\(_{1.19}\) & 73.54\(_{4.13}\) & 12.22\(_{21.17}\) & 99.29\(_{0.29}\) & 99.94\(_{0.02}\) & \underline{99.97\(_{0.01}\)} & \textbf{100.00\(_{0.00}\)} \\
    \midrule

    \multirow{3}{*}{Magic}
    & Shape & 89.70\(_{2.19}\) & 76.87\(_{0.83}\) & 91.43\(_{0.37}\) & 87.74\(_{0.26}\) & 97.77\(_{0.19}\) & 92.31\(_{0.11}\) & \underline{98.64\(_{0.20}\)} & \textbf{98.76\(_{0.42}\)} \\
    & Trend & 89.38\(_{0.99}\) & 88.30\(_{0.39}\) & 93.87\(_{0.63}\) & 83.17\(_{0.03}\) & \textbf{99.17\(_{0.23}\)} & 91.10\(_{0.05}\) & 98.26\(_{0.22}\) & \underline{98.73\(_{0.54}\)} \\
    & SA & 87.75\(_{0.99}\) & 97.39\(_{0.80}\) & 98.56\(_{0.22}\) & 44.47\(_{0.06}\) & 99.27\(_{0.11}\) & 93.02\(_{0.05}\) & \underline{99.28\(_{0.06}\)} & \textbf{99.74\(_{0.05}\)} \\

        \midrule
    \multirow{3}{*}{Ave.}
    & Shape & 83.58 & 75.77 & 84.07 & 89.97 & 78.28 & 93.43 & \underline{98.03} & \textbf{98.29} \\
    & Trend & 85.21 & 81.73 & 86.10 & 67.60 & 78.66 & 88.74 & \underline{97.32} & \textbf{98.05} \\
    & SA & 28.39 & 76.60 & 33.80 & 17.64 & 80.97 & 75.20 & \underline{81.38} & \textbf{82.23} \\
    
    \bottomrule
    \end{tabular}
  }
  \label{tab:data_fidelity}
  \vspace{-10pt}
\end{table*}

$\rhd$ \textbf{\ours\ achieves strong distributional fidelity.}
As shown in Table~\ref{tab:data_fidelity}, \ours\ obtains the best Shape score on 5 out of 6 datasets, with an average Shape of 98.29\%, compared with 98.03\% for the strongest diffusion baseline TABDIFF. The improvements are consistent in Beijing, News, Adult, Default, and Magic. For example, \ours\ improves Shape from 98.35\% to 98.56\% on Beijing and from 98.84\% to 99.06\% on Adult. These results indicate that column-wise diffusion in the unified semantic space preserves marginal column distributions effectively.

$\rhd$ \textbf{\ours\ improves semantic fidelity and dependency preservation.}
For Trend, \ours\ achieves the best result on 5 out of 6 datasets, with an average score of 98.05\%, outperforming TABDIFF by 0.73\%percentage points on average. The gain is especially clear on Default, where \ours\ improves Trend from 96.46\% to 97.89\%. For SA, \ours\ also achieves the best result on 5 out of 6 datasets, improving the average score from 81.38\% for TABDIFF to 82.23\%. Notably, \ours\ reaches 99.98\% on Shoppers, 93.76\% on Adult, 100.00\% on Default, and 99.74\% on Magic. These results suggest that conditioning score estimation on \(\mathbf P\), \(e_i\), and \(h_{t,-i}\) helps generate rows that are not only distributionally plausible but also semantically consistent.

\vspace{-5pt}
\subsubsection{Task Fidelity}
\vspace{-5pt}
\label{sub:task_fidelity}

To answer Q3, we evaluate downstream utility by training predictive models on synthetic data and testing them on real data. Regression datasets are evaluated by RMSE, while classification datasets are evaluated by AUC, as reported in Table~\ref{tab:mle}.

\begin{table*}[t]
  \centering
  \caption{MLE scores to evaluate downstream task availability. The best results are highlighted in bold, and suboptimal ones are marked with an underline.}
  \resizebox{0.95\textwidth}{!}{
    \begin{tabular}{c|l|c|ccccccc|c}
    \toprule
    \multirow{2}[4]{*}{Data} & \multirow{2}[4]{*}{Metrics} & \multirow{2}[4]{*}{Real} & \multicolumn{8}{c|}{Method} \multirow{2}[4]{*}{\textbf{\ours}}\\
\cmidrule{4-10}          &       &       & CTGAN & CTGAN+ & TVAE  & P-TA  & TabDDPM & TABSYN & TABDIFF &  \\
 \midrule
    Beijing & RMSE$\downarrow$   & 0.421\(_{0.005}\) & 0.850\(_{0.054}\) & 0.960\(_{0.180}\) & 0.833\(_{0.058}\) & 1.324\(_{0.131}\) & 0.634\(_{1.331}\) & 0.613\(_{0.024}\) & \underline{0.582\(_{0.012}\)} & \textbf{0.498\(_{0.005}\)} \\
    News & RMSE$\downarrow$   & 0.853\(_{0.005}\) & 0.860\(_{0.015}\) & 6.840\(_{0.030}\) & 0.993\(_{0.045}\) & \underline{0.821\(_{0.011}\)} & 3.104\(_{0.000}\) & 0.846\(_{0.019}\) & 0.864\(_{0.021}\) & \textbf{0.817\(_{0.003}\)} \\
    Shoppers & AUC$\uparrow$   & 0.924\(_{0.100}\) & 0.880\(_{0.740}\) & 0.881\(_{1.412}\) & 0.885\(_{2.024}\) & 0.913\(_{0.543}\) & 0.814\(_{18.364}\) & 0.888\(_{0.763}\) & \underline{0.923\(_{0.214}\)} & \textbf{0.932\(_{0.201}\)} \\
    Adult & AUC$\uparrow$   & 0.925\(_{0.250}\) & 0.892\(_{0.132}\) & 0.898\(_{0.073}\) & 0.887\(_{1.194}\) & \textbf{0.916\(_{0.152}\)} & 0.909\(_{0.380}\) & 0.897\(_{0.432}\) & 0.912\(_{0.081}\) & \underline{0.913\(_{0.072}\)} \\
    Default & AUC$\uparrow$   & 0.765\(_{0.133}\) & 0.735\(_{0.744}\) & 0.687\(_{7.472}\) & 0.726\(_{1.362}\) & 0.743\(_{0.170}\) & \underline{0.759\(_{0.182}\)} & 0.750\(_{0.784}\) & 0.746\(_{0.392}\) & \textbf{0.771\(_{0.481}\)} \\
    Magic & AUC$\uparrow$   & 0.947\(_{0.213}\) & 0.830\(_{0.614}\) & 0.873\(_{0.451}\) & 0.901\(_{0.470}\) & 0.885\(_{0.593}\) & 0.932\(_{0.552}\) & 0.854\(_{0.201}\) & \underline{0.936\(_{0.430}\)} & \textbf{0.937\(_{0.402}\)} \\
    \bottomrule
    \end{tabular}%
    }
  \label{tab:mle}%
  \vspace{-10pt}
\end{table*}%

$\rhd$ \textbf{\ours\ generates task-useful synthetic data.}
As shown in Table~\ref{tab:mle}, \ours\ achieves the best downstream performance on 5 out of 6 datasets and the second-best result on Adult. On the two regression datasets, \ours\ obtains the lowest RMSE, reducing the error from 0.582 to 0.498 on Beijing compared with the strongest baseline. For classification tasks, \ours\ achieves the highest AUC on Shoppers, Default, and Magic, respectively.
These results indicate that \ours\ does not only improve distributional and semantic fidelity, but also preserves task-relevant information for downstream learning. In particular, the strong performance on both regression and classification datasets suggests that prior-conditioned denoising helps maintain predictive column semantics and cross-column dependency patterns, thereby providing more reliable synthetic supervision.

\vspace{-5pt}
\subsection{Analysis of Weak Semantic Priors}
\vspace{-5pt}
\label{sub:prior_analysis}

\begin{wrapfigure}{r}{0.48\textwidth}
    \centering
    \vspace{-10pt}
    \includegraphics[width=0.46\textwidth]{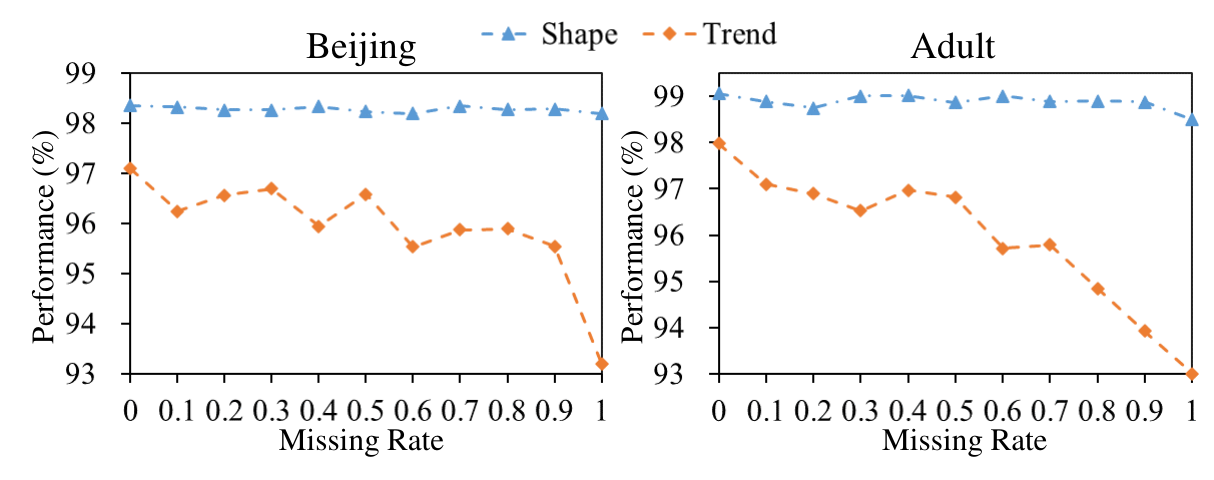}
    \vspace{-10pt}
    \caption{\label{figure:robustness}Robustness to incomplete semantic priors on Beijing and Adult.}
    \vspace{-10pt}
\end{wrapfigure}

To answer Q4, we evaluate the robustness of \ours\ under incomplete semantic priors. Since semantic priors are often weakly specified and may be only partially available, we randomly remove different proportions of priors and report Shape and Trend on Beijing and Adult in Fig.~\ref{figure:robustness}.

$\rhd$\textbf{\ours\ remains stable under moderate prior missingness.}
As shown in Fig.~\ref{figure:robustness}, Shape remains highly stable across different missing rates. On Beijing, Shape stays around 98\% even when the missing rate increases to 1.0. On Adult, Shape remains close to 99\% under most missing ratios. This indicates that marginal distribution modeling is mainly supported by column-wise diffusion in the unified semantic space and is less sensitive to prior availability.

$\rhd$\textbf{Semantic priors mainly benefit dependency preservation.}
Compared with Shape, Trend is more sensitive to missing priors. On Beijing, Trend remains relatively stable under moderate missing rates but drops sharply when almost all priors are removed. On Adult, Trend gradually decreases from around 98\% to nearly 93\% as the missing rate increases. These results show that weak semantic priors mainly contribute to cross-column dependency modeling, while \ours\ can still operate effectively when priors are only partially available.

\subsection{Ablation Study}
\label{sub:ablation}

To evaluate the contribution of each key design in \ours, we conduct ablation studies in Table~\ref{tab:abl}. 
USS denotes the unified semantic space that represents tabular values, column identities, and semantic priors in a shared latent space. 
ICS denotes intra-column semantics, which provide attribute-level plausibility constraints. 
ISR denotes inter-column symbolic rules, which provide tuple-level dependency constraints. 
Cw-FP denotes the column-wise forward process, which performs column-level corruption in the semantic latent space.

\begin{wraptable}{r}{0.41\textwidth}
\vspace{-10pt}
\centering 
\captionsetup{width=0.41\textwidth}
\caption{Ablation results of key designs in \ours on Default.} 

\resizebox{0.41\textwidth}{!}{ 
\begin{tabular}{cccc|ccc} 
\toprule 
\multicolumn{4}{c|}{Key Components} 
& \multicolumn{3}{c}{Metrics} \\ 
\cmidrule{1-7} 
USS & ICS & ISR & Cw-FP & AUC & Shape & Trend \\ 
\midrule 
$\times$ & $\times$ & $\times$ & $\times$ 
& 0.741\(_{0.553}\) & 96.84\(_{0.32}\) & 93.98\(_{0.49}\) \\ 

$\checkmark$ & $\times$ & $\times$ & $\times$ 
& 0.752\(_{0.523}\) & 97.18\(_{0.35}\) & 95.21\(_{0.48}\) \\ 

$\checkmark$ & $\checkmark$ & $\times$ & $\times$ 
& 0.763\(_{0.483}\) & 97.56\(_{0.31}\) & 95.94\(_{0.55}\) \\ 

$\checkmark$ & $\checkmark$ & $\checkmark$ & $\times$ 
& 0.769\(_{0.442}\) & 97.79\(_{0.14}\) & 97.95\(_{0.22}\) \\ 

$\checkmark$ & $\checkmark$ & $\checkmark$ & $\checkmark$ 
& 0.771\(_{0.481}\) & 98.57\(_{0.20}\) & 97.89\(_{0.30}\) \\ 
\bottomrule 
\end{tabular}%
} 
\label{tab:abl}%
\vspace{-10pt}
\end{wraptable}

$\rhd$\textbf{Semantic conditions improve both utility and fidelity.}
Introducing USS improves AUC from 0.741 to 0.752, Shape from 96.84\% to 97.18\%, and Trend from 93.98\% to 95.21\%, showing that a unified semantic space provides a useful foundation for heterogeneous tabular generation. Adding ICS further improves all metrics, indicating that intra-column semantics help preserve attribute-level plausibility. Incorporating ISR further raises Trend from 95.94\% to 97.95\%, which confirms that inter-column symbolic rules mainly benefit cross-column dependency modeling.

$\rhd$\textbf{Column-wise forward process enhances marginal fidelity.}
With Cw-FP, \ours\ achieves the best AUC and Shape scores, 0.771 and 98.57\%, respectively, while maintaining a strong Trend score of 97.89\%. This suggests that column-wise corruption improves marginal distribution preservation without undermining semantic dependency modeling.

\subsection{Visualization of Fine-Grained Tabular Fidelity}
\label{sub:cat_vis}

\begin{wrapfigure}{r}{0.50\textwidth}
    \centering
    \vspace{-10pt}
    \includegraphics[width=0.48\textwidth]{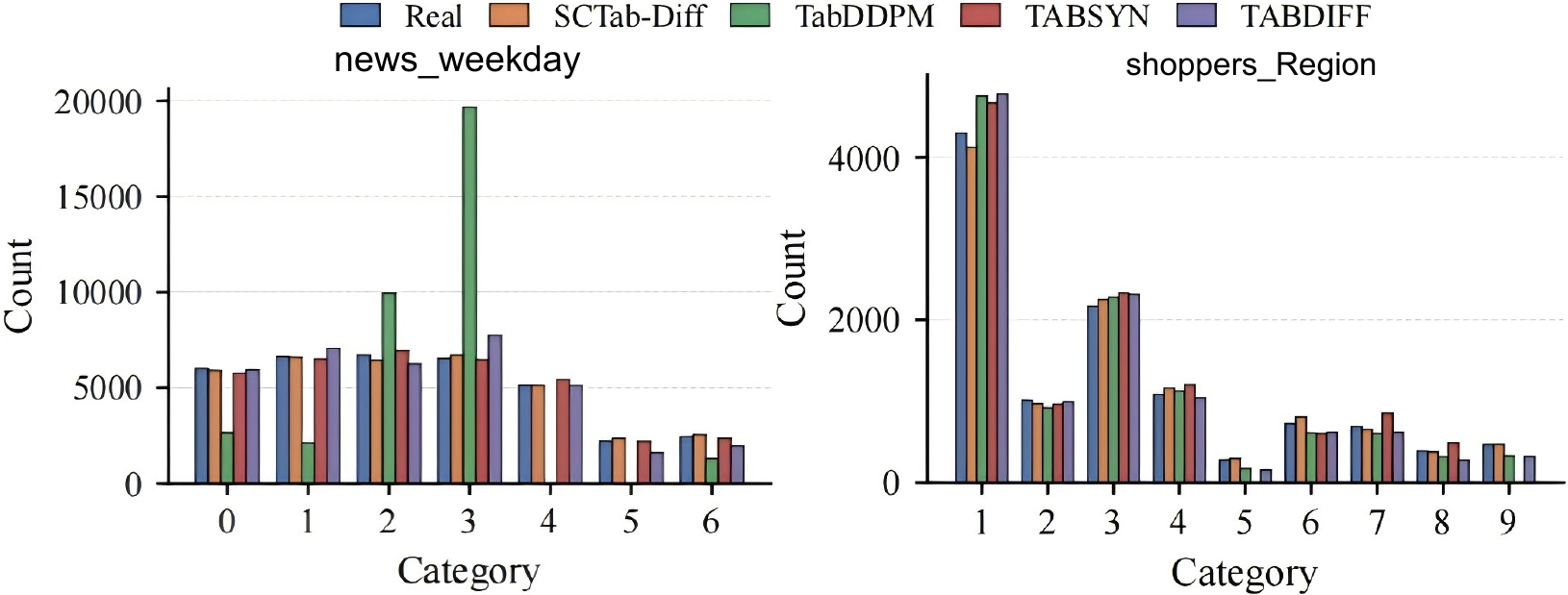}
    \vspace{-8pt}
    \caption{\label{figure:cat_fidelity}Visualization of categorical column fidelity on News and Shoppers.}
    \vspace{-10pt}
\end{wrapfigure}

To further examine fine-grained tabular fidelity, we visualize representative categorical columns in Fig.~\ref{figure:cat_fidelity}. Specifically, we compare the category-wise sample counts of \emph{Weekday} in News and \emph{Region} in Shoppers across real data and synthetic data generated by different methods. This visualization complements the Shape metric by showing whether each method preserves detailed category-level patterns rather than only aggregate column-wise fidelity.

$\rhd$\textbf{\ours\ better aligns category-wise sample counts.}
As shown in Fig.~\ref{figure:cat_fidelity}, TabDDPM exhibits clear deviations from the real data. On the \emph{Weekday} column in News, it substantially overestimates categories 2 and 3 while underestimating several other categories. TABSYN and TABDIFF reduce some deviations, but still show mismatches in specific categories. In contrast, \ours\ follows the real count pattern more closely across most categories, indicating that the unified semantic space and column-wise denoising help preserve categorical column identity during generation.

$\rhd$\textbf{\ours\ preserves minority categorical patterns.}
On the \emph{Region} column in Shoppers, \ours\ better matches the real counts for minority categories such as 5, 8, and 9, while several baselines show larger deviations. This suggests that \ours\ does not only recover dominant categories, but also preserves low-frequency categorical patterns. Such behavior is important for tabular generation because minority categories often carry useful downstream information.

$\rhd$\textbf{Additional numerical and inter-column visualizations.}
Due to space limitations, additional visualizations are provided in Fig.~\ref{figure:vis_exm2} and Fig.~\ref{figure:vis_exm3} of Appendix~\ref{suba:vis_exm2} and Appendix~\ref{suba:vis_exm3}. Fig.~\ref{figure:vis_exm2} shows that \ours\ better fits representative numerical columns such as \emph{BounceRates} in Shoppers and \emph{n\_tokens\_title} in News, while TabDDPM, TABSYN, and TABDIFF exhibit visible shifts or mismatches around high-density regions. Fig.~\ref{figure:vis_exm3} further shows that \ours\ preserves pair-wise dependency patterns more faithfully than the baselines.

\vspace{-5pt}
\section{Conclusion}
\vspace{-5pt}
This paper introduces \ours, a semantics-consistent diffusion framework for tabular data generation under weakly specified semantic priors. The key idea is to move beyond pure distribution matching by treating intra-column semantics and inter-column symbolic rules as generation conditions during reverse denoising. To this end, \ours\ aligns heterogeneous column values, column identities, and semantic priors in a unified semantic space, and performs prior-conditioned denoising to preserve both column-wise distributions and cross-column semantic dependencies. Experiments on six real-world benchmark datasets demonstrate that \ours\ improves distributional fidelity, semantic consistency, and downstream task utility compared with representative tabular generation baselines. Additional analyses under incomplete priors further show that the proposed framework can remain robust when semantic knowledge is only partially available.


\medskip

{\small
\bibliographystyle{unsrt}
\bibliography{neurips_2026}
}
\newpage
\appendix
\section{Related Work}

\subsection{Distribution-Centric Tabular Generation}

Early studies on synthetic tabular data mainly follow a distribution-centric paradigm, where the goal is to approximate the empirical distribution of real tables. GAN-based methods are widely used in this line. CTGAN~\cite{xu2019modeling} introduces mode-specific normalization to better handle non-Gaussian continuous columns and conditional generation to address imbalanced categorical distributions. CTAB-GAN+~\cite{zhao2024ctab} further improves the stability and utility of GAN-based tabular synthesis. Other representative methods include TableGAN~\cite{park2018data}, which incorporates label information into the generation process, and MedGAN~\cite{choi2017generating}, which targets discrete medical records. In parallel, VAE-based models such as TVAE~\cite{xu2019modeling} and VAEM~\cite{ma2020vaem} have been developed to model heterogeneous tabular variables through latent representations. GOGGLE~\cite{liu2023goggle} further introduces relational structure learning to capture dependencies among features.

Recent work also explores the use of large language models for tabular data synthesis and augmentation. For example, P-TA~\cite{yang2024p} uses proximal policy optimization to guide LLM-based tabular augmentation. EPIC~\cite{kim2024epic} studies effective prompting for imbalanced tabular classification. Curated LLM~\cite{seedat2023curated} further investigates the synergy between LLMs and data curation in low-data regimes. Although these methods improve tabular synthesis from different perspectives, they still mainly optimize distributional or task-level fidelity. They rarely model explicit semantic validity constraints that govern whether a generated row is logically plausible.

\vspace{-6pt}

\subsection{Diffusion Models for Mixed-Type Tabular Data}

Diffusion models have recently become a strong paradigm for tabular generation due to their stable training and strong distribution modeling ability. Since tabular data contains both numerical and categorical columns, early diffusion-based methods often design separate mechanisms for different data types. TabDDPM~\cite{kotelnikov2023tabddpm} applies Gaussian diffusion to numerical features and multinomial diffusion to categorical features. CoDi~\cite{lee2023codi} further introduces co-evolving diffusion processes for continuous and discrete variables. TabDDPM-HER~\cite{ceritli2023synthesizing} adapts diffusion-based tabular generation to heterogeneous electronic health records.

Another line of work performs diffusion in latent space. TABSYN~\cite{zhangmixed} first maps mixed-type tabular data into a continuous latent space and then applies score-based diffusion for synthesis. FLEXGEN-EHR~\cite{he2024flexible} incorporates optimal transport to align heterogeneous EHR features in latent space. CTSyn~\cite{lin2024ctsyn} trains across multiple datasets and aims to build a foundation model for cross-table generation. More recently, TABDIFF~\cite{shi2025tabdiff} proposes a mixed-type diffusion model with feature-wise learnable diffusion processes, which improves the modeling of heterogeneous feature distributions.

Despite these advances, existing tabular diffusion models are still largely distribution-driven. They mainly learn statistical regularities from observed samples and recover noisy tabular representations toward high-density regions of the data distribution. However, semantic validity in tabular data is often governed by column meanings and cross-column symbolic rules. Such constraints are difficult to recover from implicit correlations alone. As a result, existing diffusion models may generate samples that are statistically plausible but semantically inconsistent.

\vspace{-6pt}

\subsection{Semantics-Aware Tabular Generation}

Realistic tabular generation requires more than matching marginal and joint distributions. A valid tabular row should also satisfy attribute-level semantic plausibility and tuple-level dependency constraints~\cite{yu2025shap}. For example, a column may have a meaningful value range, an ordinal relation, or a domain-specific interpretation. Meanwhile, multiple columns may be linked by symbolic rules that define valid co-occurrence patterns. Some recent studies have started to use external knowledge or LLMs to improve tabular data generation and augmentation~\cite{KwokWC25,yang2024p}. However, these methods mainly use language models as data generators~\cite{spinaci2026contexttab,jacob2026tabscm}, prompt-based augmenters~\cite{long2025llm,zhangetal2025}, or distribution refiners~\cite{arazi2026tabstar,yang2026sage}. They do not explicitly incorporate weak semantic priors into the generative dynamics of diffusion models. Existing tabular diffusion methods also lack a mechanism to condition denoising on column identities, intra-column semantics, and inter-column symbolic rules at the same time.
In contrast, \ours\ studies semantics-consistent tabular diffusion under weakly specified semantic priors. Instead of treating semantic constraints as post-hoc filters, \ours\ extracts intra-column semantics and inter-column symbolic rules from tabular schema and textual descriptions, then injects them into prior-conditioned denoising. This design enables the generator to preserve distributional fidelity while reducing rule-violating and semantically implausible synthetic rows.

\section{Algorithm and Theoretical Analysis}
\subsection{Algorithms}
\label{suba:alg}

\begin{algorithm}[t]
\caption{Training of \ours}
\label{alg:train}
\begin{algorithmic}[1]
\Require Tabular dataset \(D=\{X,F\}\), weak semantic priors \(P\), masking probability \(p_u\), noise schedules \(\{\sigma_{\rho_i}(t)\}_{i=1}^{n}\)
\Ensure Trained denoising network \(s_\theta\) and tabular decoder
\State Encode semantic priors into the unified semantic space: \(\mathbf P=\mathrm{PLM}(P)\)
\Repeat
    \State Sample a mini-batch \(x\sim X\) and timestep \(t\sim p(t)\)
    \State Sample prior mask \(M=\mathbf{1}\{U>p_u\},\ U\sim\mathrm{Uniform}(0,1)^m\)
    \For{\(i=1,\dots,n\)}
        \State Construct column representation \(h_{0,i}\) and column identity embedding \(e_i\) by Eq.~\eqref{eq:semantic_encoding} and Eq.~\eqref{eq:column_identity}
        \State Sample noise \(\epsilon_i\sim\mathcal N(0,I)\)
        \State Apply column-wise forward corruption:
        \[
        h_{t,i}=h_{0,i}+\sigma_{\rho_i}(t)\epsilon_i
        \]
        \State Annotate noisy representation with column identity:
        \[
        \tilde h_{t,i}=[h_{t,i}\|e_i]
        \]
        \State Inject semantic priors through cross-attention:
        \[
        \bar h_{t,i}=
        \mathrm{PriorInjection}(\tilde h_{t,i},\mathbf P*M)
        \]
    \EndFor
    \State Capture cross-column dependencies:
    \[
    \hat h_{t,1},\dots,\hat h_{t,n}
    =
    \mathrm{DepCapture}(\bar h_{t,1},\dots,\bar h_{t,n})
    \]
    \For{\(i=1,\dots,n\)}
        \State Predict noise: \(\hat\epsilon_i=\mathrm{MLP}(\hat h_{t,i})\)
    \EndFor
    \State Compute diffusion loss Eq.~\eqref{eq:diff_loss}:
    \[
    \mathcal L_{\mathrm{diff}}
    =
    \sum_{i=1}^{n}\|\hat\epsilon_i-\epsilon_i\|_2^2
    \]
    \State Decode clean latent representations \(h_{0,1},\dots,h_{0,n}\) and compute reconstruction loss Eq.~\eqref{eq:rec_loss}:
    \[
    \mathcal L_{\mathrm{rec}}
    =
    \ell_{\mathrm{recon}}(x,\hat x)
    \]
    \State Update parameters by minimizing
    \[
    \mathcal L=\mathcal L_{\mathrm{diff}}+\lambda\mathcal L_{\mathrm{rec}}
    \]
\Until{convergence}
\end{algorithmic}
\end{algorithm}

\begin{algorithm}[t]
\caption{Synthetic Tabular Data Generation of \ours}
\label{alg:sample}
\begin{algorithmic}[1]
\Require Trained \ours, column names \(F=\{f_i\}_{i=1}^{n}\), weak semantic priors \(P\), number of synthetic samples \(N_{\mathrm{syn}}\)
\Ensure Synthetic tabular dataset \(\hat D\)
\State Encode semantic priors: \(\mathbf P=\mathrm{PLM}(P)\)
\State Encode column identities: \(e_i=\mathrm{Average}(\mathrm{PLM}(f_i)),\ i=1,\dots,n\)
\State Set \(M=\mathbf{1}\) unless sampling under incomplete-prior settings
\For{\(m=1,\dots,N_{\mathrm{syn}}\)}
    \For{\(i=1,\dots,n\)}
        \State Initialize latent variable Eq.~\eqref{eq:latent_forward}:
        \[
        h_{T,i}\sim\mathcal N(0,\sigma_{\rho_i}^{2}(T)I)
        \]
    \EndFor
    \For{\(t=T,\dots,1\)}
        \For{\(i=1,\dots,n\)}
            \State Annotate noisy latent representation:
            \[
            \tilde h_{t,i}=[h_{t,i}\|e_i]
            \]
            \State Inject semantic priors Eq.~\eqref{eq:prior_injection}:
            \[
            \bar h_{t,i}=
            \mathrm{PriorInjection}(\tilde h_{t,i},\mathbf P*M)
            \]
        \EndFor
        \State Capture cross-column dependencies:
        \[
        \hat h_{t,1},\dots,\hat h_{t,n}
        =
        \mathrm{DepCapture}(\bar h_{t,1},\dots,\bar h_{t,n})
        \]
        \For{\(i=1,\dots,n\)}
            \State Predict noise: \(\hat\epsilon_i=\mathrm{MLP}(\hat h_{t,i})\)
            \State Approximate the prior-conditioned score:
            \[
            s_\theta(h_{t,i},t,\mathbf P,e_i,h_{t,-i})
            =
            -\frac{\hat\epsilon_i}{\sigma_{\rho_i}(t)}
            \]
            \State Update \(h_{t,i}\) to \(h_{t-1,i}\) using the reverse denoising step in Eq.~\eqref{eq:re_new}
        \EndFor
    \EndFor
    \State Obtain denoised latent representations \(\hat h_{0,1},\dots,\hat h_{0,n}\)
    \For{\(i=1,\dots,n\)}
        \State Decode \(\hat h_{0,i}\) into \(\hat x_i\) using the tabular decoder in Eq.~\eqref{eq:decoder}
    \EndFor
    \State Form one synthetic row \(\hat x=(\hat x_1,\dots,\hat x_n)\) and add it to \(\hat D\)
\EndFor
\State \Return \(\hat D\)
\end{algorithmic}
\end{algorithm}
In this section, we summarize the training and synthetic data generation procedures of \ours. Algorithm~\ref{alg:train} presents the training process, while Algorithm~\ref{alg:sample} describes the reverse denoising and reconstruction process used for synthetic tabular data generation.

During training, \ours\ first encodes the weak semantic priors \(P\) into the unified semantic space as \(\mathbf P=\mathrm{PLM}(P)\). For each mini-batch, a timestep \(t\) and a prior mask \(M\) are sampled. Each column value \(x_i\) is then mapped into its initial semantic representation \(h_{0,i}\), and the corresponding column identity \(f_i\) is encoded as \(e_i\). Gaussian noise is injected into each column representation through the column-wise forward process, yielding \(h_{t,i}=h_{0,i}+\sigma_{\rho_i}(t)\epsilon_i\). To implement prior-conditioned denoising, the noisy representation is first annotated with its column identity, and the masked semantic prior representation \(\mathbf P*M\) is then injected through cross-attention. The resulting prior-enhanced column representations are passed into the dependency capture module to model cross-column context. Finally, the denoising network predicts the injected noise \(\hat{\epsilon}_i\), and the diffusion loss is computed by matching \(\hat{\epsilon}_i\) with the true noise \(\epsilon_i\). The decoder is trained with the reconstruction objective, and all trainable parameters are optimized by minimizing the combined training loss.

During generation, \ours\ starts from Gaussian latent variables \(h_{T,i}\) sampled independently for each column. At each reverse denoising step, the model uses the column identity embedding \(e_i\), the semantic prior embedding \(\mathbf P\), and the current cross-column context \(h_{t,-i}\) to approximate the prior-conditioned score. The reverse process iteratively updates \(h_{t,i}\) to \(h_{t-1,i}\) until denoised latent representations \(\hat h_{0,1},\dots,\hat h_{0,n}\) are obtained. These representations are then decoded column by column through the tabular decoder to form a synthetic row \(\hat{x}\). Repeating this process yields the synthetic tabular dataset \(\hat D\).

It is worth noting that the PLM encoder and the tabular reconstruction module follow the pretraining strategy of~\cite{zhangmixed}. Their parameters are kept frozen when training the denoising network of \ours.

\subsection{Analysis of Prior-Conditioned Denoising}
\label{app:theory_analysis}

This section provides an interpretation of why prior-conditioned denoising can improve semantic consistency. The analysis does not introduce an additional training objective. Instead, it explains how weak semantic priors change the reverse denoising direction.

\paragraph{Bayesian view of semantic conditioning.}
Recall that \(\mathcal{V}(P)\) denotes the semantic validity set induced by weak semantic priors \(P\). We define the semantic validity event as
\begin{equation}
\mathcal{E}_{P}=\{x_0\in\mathcal{V}(P)\},
\end{equation}
which means that the clean tabular row satisfies the weak semantic priors. In conventional diffusion, the reverse process follows the distributional score \(\nabla_{h_t}\log p_t(h_t)\), which guides noisy samples toward high-density regions of the perturbed data distribution. In contrast, semantics-consistent generation should follow the score of the perturbed distribution conditioned on the validity event:
\begin{equation}
\nabla_{h_t}\log p_t(h_t\mid \mathcal{E}_{P}).
\end{equation}
By Bayes' rule, we have
\begin{equation}
p_t(h_t\mid \mathcal{E}_{P})
=
\frac{
p_t(h_t)\Pr(\mathcal{E}_{P}\mid h_t)
}{
\Pr(\mathcal{E}_{P})
}.
\end{equation}
Taking the gradient with respect to \(h_t\) gives
\begin{equation}
\nabla_{h_t}\log p_t(h_t\mid \mathcal{E}_{P})
=
\nabla_{h_t}\log p_t(h_t)
+
\nabla_{h_t}\log \Pr(\mathcal{E}_{P}\mid h_t).
\label{eq:semantic_score_decomp}
\end{equation}
Eq.~\eqref{eq:semantic_score_decomp} shows that the semantics-aware score can be decomposed into two terms. The first term is the standard distributional score, which encourages statistical realism. The second term increases the probability that the noisy representation will be denoised into a semantically valid row. Therefore, semantic priors affect the denoising direction before a complete sample is generated, rather than being applied as post-hoc filters.

\paragraph{Connection to prior-conditioned score estimation.}
In \ours, we do not explicitly estimate \(\Pr(\mathcal{E}_{P}\mid h_t)\). Instead, the denoising network amortizes this effect by conditioning score estimation on weak semantic priors, column identity, and cross-column context:
\begin{equation}
s_\theta(h_{t,i},t,\mathbf{P},e_i,h_{t,-i})
\approx
-\sigma_{\rho_i}(t)
\nabla_{h_{t,i}}
\log p_t(h_{t,i}\mid \mathbf{P},e_i,h_{t,-i}).
\end{equation}
Here, \(\mathbf{P}\) provides semantic prior information, \(e_i\) specifies the column identity, and \(h_{t,-i}\) provides row-level context. This conditional score can be viewed as an amortized approximation to the semantics-aware score in Eq.~\eqref{eq:semantic_score_decomp}. In this sense, the model learns to move noisy column representations toward regions that are not only likely under the data distribution, but also compatible with weak semantic priors.

\paragraph{Effect on invalid regions.}
The decomposition above also explains why prior-conditioned denoising can reduce invalid samples. For any noisy latent representation \(h_t\), Bayes' rule gives
\begin{equation}
\frac{p_t(h_t\mid \mathcal{E}_{P})}{p_t(h_t)}
=
\frac{\Pr(\mathcal{E}_{P}\mid h_t)}{\Pr(\mathcal{E}_{P})}.
\label{eq:semantic_reweight}
\end{equation}
Thus, noisy states that are more likely to denoise into semantically valid rows receive larger relative density under the semantics-conditioned distribution. For two noisy states \(h_t^{a}\) and \(h_t^{b}\), if
\begin{equation}
\Pr(\mathcal{E}_{P}\mid h_t^{a})
>
\Pr(\mathcal{E}_{P}\mid h_t^{b}),
\end{equation}
then
\begin{equation}
\frac{
p_t(h_t^{a}\mid \mathcal{E}_{P})/
p_t(h_t^{b}\mid \mathcal{E}_{P})
}{
p_t(h_t^{a})/
p_t(h_t^{b})
}
=
\frac{
\Pr(\mathcal{E}_{P}\mid h_t^{a})
}{
\Pr(\mathcal{E}_{P}\mid h_t^{b})
}
>1.
\label{eq:relative_validity}
\end{equation}
This means that semantic conditioning increases the relative likelihood of trajectories that can lead to valid rows. In contrast, trajectories that are likely to end in \(\mathcal{V}^{c}(P)\), the complement of the validity set, are relatively down-weighted. This provides an intuitive explanation for why \ours\ can reduce tuple-level semantic violations while maintaining distributional fidelity.

\paragraph{Robustness induced by prior masking.}
Weak semantic priors may be incomplete in real applications. To improve robustness, \ours\ randomly masks prior tokens during training. Let \(M\in\{0,1\}^{m}\) denote the prior mask, where each prior token is retained with probability \(1-p_u\). The training objective can be viewed as minimizing the expected denoising loss over different available prior subsets:
\begin{equation}
\mathcal{L}_{\mathrm{diff}}
=
\mathbb{E}_{t,\epsilon,M}
\sum_{i=1}^{n}
\left\|
\epsilon_i
-
\epsilon_\theta(h_{t,i},t,\mathbf{P}\ast M,e_i,h_{t,-i})
\right\|_2^2.
\label{eq:masked_objective}
\end{equation}
The expected number of active prior tokens is:
\begin{equation}
\mathbb{E}\left[\|M\|_0\right]=(1-p_u)m.
\end{equation}
Therefore, prior masking trains the model under multiple partial-prior conditions rather than assuming complete prior availability. This acts as a regularizer and prevents the denoising network from over-relying on a fixed, complete prior set. A larger \(p_u\) weakens semantic guidance, while a moderate \(p_u\) improves robustness to incomplete priors.

\section{Detailed Experiment Setups}
\subsection{Datasets}
\label{suba:datasets}
\begin{table*}[htbp]
  \centering
  \caption{Statistics of datasets. \# Num/Cat stands for the number of numerical columns and the number of categorical columns, respectively. \#  Weak Semantic Priors for the number of Inter-Column Symbolic Rules and Intra-Column Semantics, respectively.}
  \resizebox{1.0\textwidth}{!}{
    \begin{tabular}{cccccc}
    \toprule
    Dataset & \# Rows & \# Num/Cat & \# Train/Validation/Test & \# Weak Semantic Priors & Task \\
    \midrule
    Adult & 48842 & 6/9   & 28943/3618/16281 & 3/13  & Classification \\
    Default & 30000 & 14/11 & 24000/3000/3000 & 3/24  & Classification \\
    Shoppers & 12330 & 10/8  & 9864/1233/1233 & 5/17  & Classification \\
    Magic & 19019 & 10/1  & 15215/1902/1902 & 5/10  & Classification \\
    Beijing & 43824 & 7/5   & 35058/4383/4383 & 5/11  & Regression \\
    News  & 39644 & 46/2  & 31714/3965/3965 & 10/47 & Regression \\
    \bottomrule
    \end{tabular}%
    }
  \label{tab:data}%
\end{table*}%

We use six tabular datasets from the UCI machine learning library\footnote{\url{https://archive.ics.uci.edu/datasets}}, including Adult, Default, Shoppers, and Magic datasets for classification tasks, as well as Beijing and News datasets for regression tasks. The statistical information of the dataset is shown in Table~\ref{tab:data}. Weak Semantic Priors for each dataset are shown in Tables [\ref{tab:adult}-\ref{tab:news}].

\subsection{LLM-based Prior Construction and Validation}
\label{app:llm_prior_quality}

To construct weak semantic priors used by \ours, we adopt a metadata-driven LLM-assisted protocol based on \textbf{GPT-4o mini}. The LLM is provided only with dataset metadata, tabular schema, and textual descriptions, and does not access raw tabular records during candidate constraint generation. The goal is not to use the LLM as an unconstrained rule generator, but to extract candidate constraints from weak semantic sources and validate them with observed training data before they are used as semantic conditions.

\paragraph{Candidate constraint generation.}
For each dataset, we provide the LLM with the tabular schema and textual descriptions from the dataset metadata. The LLM does not access raw table records during this stage. It is instructed to generate two types of candidate priors: intra-column semantics and inter-column symbolic rules. Intra-column semantics describe the valid value space or semantic type of an individual column, such as numerical ranges, categorical domains, non-negativity, or ordinal meanings. Inter-column symbolic rules describe dependency constraints across multiple columns, such as compatibility relations, temporal consistency, or order constraints. To make the generated priors verifiable, the LLM is asked to express candidate constraints as Python-style checking code.

\paragraph{Validation with observed data.}
Since LLM-generated constraints may be incomplete or noisy, we validate each candidate constraint on the real training split. For each rule $r_k$, we compute its violation rate as:
\[
v(r_k) =
\frac{1}{|\mathcal{D}_{\mathrm{train}}|}
\sum_{x \in \mathcal{D}_{\mathrm{train}}}
\mathbb{I}\left[\phi_k(x)=0\right],
\]
where $\phi_k(x)$ is the Boolean checking function of rule $r_k$. Candidate constraints with low violation rates and clear metadata support are retained as weak semantic priors. Constraints with high violation rates are discarded if they are not supported by the metadata, or revised when the violation pattern indicates that the original rule is overly strict. This validation step is performed only on the training split and does not use the test split or generated samples. The real training data is never exposed to the LLM during candidate constraint generation; it only serves as a post-generation validation source for filtering or revising unreliable constraints.

\paragraph{Natural-language verbalization.}
After validation, the retained constraints are converted into natural-language descriptions for reporting. The LLM is only allowed to verbalize the validated constraints and is not allowed to introduce new rules. The final descriptions are organized into Dataset Constraints, Intra-Column Semantics, and Inter-Column Symbolic Rules, as shown in Tables~\ref{tab:adult}--\ref{tab:news}. These validated priors are then encoded and used as semantic conditions in the prior-conditioned denoising process of \ours.

\begin{table*}[t]
\centering
\small
\caption{Prompt templates used for LLM-based weak semantic prior construction. }
\label{tab:llm_prompt_templates}
\begin{tabular}{p{0.15\textwidth} p{0.25\textwidth} p{0.50\textwidth}}
\toprule
Stage & Input & Prompt Template \\
\midrule

Constraint generation
&
Dataset metadata and column schema.
&
\textit{Given the following dataset metadata and column schema, identify semantic constraints that should be satisfied by valid tabular rows. Please extract both intra-column semantics and inter-column symbolic rules. Intra-column semantics describe valid value ranges, valid categories, non-negativity, ordinal meanings, or semantic types of individual columns. Inter-column symbolic rules describe logical dependencies between two or more columns. Please express each constraint as executable Python-style checking code. Do not introduce constraints that are not supported by the metadata or column definitions.}
\\

\midrule

Natural-language conversion
&
Validated Python-style constraints and dataset metadata.
&
\textit{Given the dataset metadata and the validated Python-style constraints, convert the constraints into concise natural-language descriptions. Organize the output into three parts: Dataset Constraints, Intra-Column Semantics, and Inter-Column Symbolic Rules. Do not introduce new constraints. Use exact column names and preserve the meaning of the validated checking code.}
\\

\bottomrule
\end{tabular}
\end{table*}

\subsection{Baselines}
\label{suba:baseline}
In this section, we introduce the baseline methods used in this paper in three groups:

1) Traditional methods include GAN-based methods (e.g., CTGAN and CTGAN+) and VAE-based methods (e.g., TVAE):
\begin{itemize}
    \item CTGAN and TVAE are two methods for synthetic tabular data generation proposed by~\cite{xu2019modeling}. While both methods share the same underlying framework, they are built on different generative models: CTGAN is based on GANs, whereas TVAE relies on VAEs. Both approaches incorporate two key components: (1) mode-specific normalization, designed to handle numerical columns with complex distributions; and (2) conditional generation of numerical columns based on categorical columns, aimed at addressing imbalance issues.
    \item CTAB-GAN+~\citep{zhao2024ctab} builds upon existing methods CTGAN by incorporating RDP-based privacy accounting, similar to DP-WGAN~\citep{huang2022dpwgan}. Furthermore, by leveraging the Was+GP loss, CTAB-GAN+ effectively constrains the gradient norm, eliminating the need for weight clipping and resulting in more stable training for differentially private GANs.
\end{itemize}
2) LLM-based methods:
\begin{itemize}
    \item P-TA~\citep{yang2024p} proposes the use of Proximal Policy Optimization (PPO~\citep{schulman2017proximal}) to apply GANs to guide LLMs to refine the probability distribution of tabular features and thereby generate more realistic data.
\end{itemize}
3) Diffusion model methods:
\begin{itemize}
    \item TabDDPM~\citep{kotelnikov2023tabddpm} addresses the difficulty of handling categorical features in diffusion models by introducing additional categorical diffusion models specifically for categorical features. Despite its simplicity, our experiments have shown that TabDDPM achieves excellent performance.
    \item TABSYN~\citep{zhangmixed} synthesizes tabular data by leveraging a diffusion model within a VAE-crafted latent space. Meanwhile, it adopts a simplified forward diffusion process, which adds Gaussian noises of linear standard deviation with respect to time, thus improving sampling speed.
    \item TABDIFF~\citep{shi2025tabdiff}  introduces a joint continuous-time diffusion process for both numerical and categorical data, enabling the modeling of multimodal distributions of tabular data within a single unified framework. To address the large differences in the distributions of individual features, TABDIFF employs a feature-based learnable diffusion process, which improves the model's ability to accurately capture the overall data distribution.

\end{itemize}

\paragraph{Discussion on constrained baseline variants.} \label{app:constrained_baselines}
Post-hoc constrained variants, such as rejection sampling or rule-based repair, are not included as primary baselines because they evaluate a different mechanism from \ours. \ours\ does not apply weak semantic priors after generation. Instead, it encodes them as semantic embeddings and injects them into the reverse denoising process as generation conditions. In contrast, post-hoc filtering may improve rule satisfaction by discarding invalid samples, but can also change the generated distribution and reduce diversity. Rule-based repair may enforce local validity, but can move samples away from the learned data manifold. Moreover, most existing tabular diffusion models do not provide a direct interface for incorporating textual or symbolic priors into denoising without architectural changes. Therefore, we focus on comparing \ours\ with representative tabular generators under the same evaluation protocol, and use ablation studies to isolate the effects of intra-column semantics, inter-column symbolic rules, the unified semantic space, and the column-wise forward process.

\subsection{Metrics}
\label{suba:metrics}
In this section, we provide a detailed introduction to all the evaluation metrics used in this paper.

\textbf{Evaluation Metrics.}~We evaluate the quality of the generated data from three aspects: \emph{distribution fidelity}, \emph{structural and semantic fidelity}, and \emph{task fidelity}. Specifically, \emph{Distribution fidelity} measures the alignment between synthetic and real data distributions. We focus on per-column marginal distributions using Shape. Additionally, we report \(\alpha\)-\textit{Precision}, \(\beta\)-\textit{Recall}, and C2ST to evaluate the overall distribution in Appendix~\ref{suba:joint} and~\ref{suba:c2st}. These metrics quantify precision, diversity/coverage, and overall distinguishability between real and synthetic samples, respectively. \emph{Structural and semantic fidelity} evaluates whether inter-column dependencies and domain-specific constraints are preserved. We use Trend to measure the preservation of relational dependencies, and SA is used to evaluate the extent to which the synthetic dataset satisfies the constraints of the real tabular data. \emph{Task fidelity} assesses the usability of the generated data in downstream tasks. We evaluate this using Machine Learning Efficiency (MLE), including RMSE and AUC as performance metrics.

\subsubsection{Shape and Trend}
\textbf{Shape.} This metric measures the column-wise density estimation performance, which includes the Kolmogorov-Smirnov Test (KST)~\citep{berger2014kolmogorov} for numerical features and the Total Variation Distance (TVD)~\citep{tao2024discriminative} for categorical data. Given two numerical distributions, $p_r(x)$ (representing real data) and $p_s(x)$ (representing synthetic data), KST quantifies the distance between these distributions by calculating the maximum discrepancy between their corresponding Cumulative Distribution Functions (CDFs):

\begin{equation}
\text{KST} = \sup_x \left| F_r(x) - F_s(x) \right|,
\end{equation}

where $F_r(x)$ and $F_s(x)$ are the CDFs of $p_r(x)$ and $p_s(x)$, respectively:

\begin{equation}
F(x) = \int_{-\infty}^x p(x) \, dx.
\end{equation}

For categorical data, the TVD is commonly used. TVD computes the frequency of each category and expresses it as a probability. The TVD score represents the average difference between the probabilities of each category:

\begin{equation}
\text{TVD} = \frac{1}{2} \sum_{\omega \in \Omega} \left| R(\omega) - S(\omega) \right|, 
\end{equation}

where $\omega$ represents all possible categories in a given column $\Omega$, and $R(\cdot)$ and $S(\cdot)$ denote the real and synthetic frequencies for these categories, respectively.

\textbf{Trend.} This metric measures the pair-wise column correlation estimation performance, which includes the Pearson Score for a pair of numerical columns and the Contingency Score for a pair of categorical columns:
\begin{equation}
\begin{aligned}
    \text{Pearson Score} = \frac{1}{2} \mathbb{E}_{x,y} \left| \rho^R(x, y) - \rho^S(x, y) \right|,\\
    \text{Contingency Score} = \frac{1}{2} \sum_{\alpha \in A} \sum_{\beta \in B} \left| R_{\alpha,\beta} - S_{\alpha,\beta} \right|,
\end{aligned}
\end{equation}
where \(\rho^R(x, y)\) and \(\rho^S(x, y)\) denotes the Pearson correlation coefficient between column \(x\) and column \(y\) of the real data and synthetic data, respectively. \(\alpha\) and \(\beta\) describe all the possible categories in column A and column B, respectively. \(R_{\alpha,\beta}\) and \(S_{\alpha,\beta}\) are the joint frequency of \(\alpha\) and \(\beta\) in the real data and synthetic data, respectively.
\subsubsection{\(\alpha\)-Precision and \(\beta\)-Recall}
Following~\cite{zhangmixed}, we adopt the \(\alpha\)-Precision and \(\beta\)-Recall metrics introduced in~\cite{alaa2022faithful} to evaluate the quality of synthetic tabular data. These two metrics are designed to provide sample-level assessments of how well the synthetic data approximates the real data distribution, focusing on two critical aspects: fidelity and coverage.

\(\alpha\)-Precision quantifies the fidelity of the synthetic data. Specifically, it measures the proportion of synthetic samples that lie within the support of the real data distribution. In essence, it evaluates whether the synthetic data points are realistic and indistinguishable from genuine data samples. A high \(\alpha\)-Precision score indicates that the synthetic generator avoids producing out-of-distribution samples. On the other hand, \(\beta\)-Recall measures the coverage of the real data by the synthetic data. It reflects how well the synthetic samples represent the entire variability of the real data. In practice, \(\beta\)-Recall evaluates whether every real data point is ``close enough'' to at least one synthetic sample, thereby assessing how comprehensively the synthetic distribution captures the diversity present in the original dataset.

Together, \(\alpha\)-Precision and \(\beta\)-Recall provide a balanced view of synthetic data quality: the former ensures fidelity, while the latter ensures completeness. These metrics are particularly important for applications such as data augmentation, privacy-preserving data sharing, and model training, where both overfitting to narrow patterns and missing important data characteristics can significantly harm downstream performance.

\subsubsection{Machine Learning Efficiency (MLE)}
To measure the ability of synthetic tabular data to support downstream task learning, we evaluate their performance using Machine Learning Efficiency (MLE). Specifically, we adopt the Training on Synthetic and Testing on Real (TSTR)~\citep{fekri2019generating} scheme. In this approach, the real dataset is first split into a training set and a testing set. Next, a generation model is used to synthesize synthetic data that matches the size of the training set. This synthetic data is then used to train an XGBoost classifier or XGBoost regressor. Finally, we evaluate these machine learning models based on the real test set, calculating the AUC scores for classification tasks and the RMSE for regression tasks, respectively.

\subsubsection{Ability to Accurately
Generate Sample}
Sample Accuracy (SA) evaluates the ability of a generative model to correctly generate samples, that is, whether the generated samples can accurately follow Weak Semantic Priors and domain knowledge, as shown in Tables [\ref{tab:adult}-\ref{tab:news}]. Enhancing the model’s ability to generate accurate and representative samples can significantly reduce the likelihood of producing biased or even toxic outputs. This improvement is particularly important for downstream tasks, where the quality and reliability of the synthetic data directly impact model performance, fairness, and safety.

\subsubsection{Detection}
Detection evaluates how difficult it is to distinguish synthetic data from real data when the two are mixed. Specifically, we adopt the Classifier Two-Sample Test (C2ST) as implemented in the SDMetrics library\footnote{\url{https://docs.sdv.dev/sdmetrics}}, where a logistic regression model serves as the discriminator. 
\section{Addition Experimental Results}
\label{suba:results}
\subsection{Joint Distribution}
\label{suba:joint}

\begin{table*}[htbp]
  \centering
  \caption{Comparison of \(\alpha\)-Precision scores. Bold Face represents the best score on each dataset and marks the suboptimal ones with an underline.}
  \resizebox{0.95\textwidth}{!}{
    \begin{tabular}{ccccccccc}
    \toprule
    Methods & Adult & Default & Shoppers & Magic & Beijing & News  & Average & Ranking \\
    \midrule
    CTGAN & 78.61\(_{1.85}\)  & 68.48\(_{0.39}\)  & 78.42\(_{3.32}\)  & 83.25\(_{0.91}\)  & 96.12\(_{1.48}\)  & 97.00\(_{1.03}\)  & 83.65  & 5 \\
    CTGAN+ & 93.35\(_{2.68}\)  & 87.57\(_{4.35}\)  & 92.73\(_{2.87}\)  & 41.74\(_{0.78}\)  & 89.27\(_{11.23}\)  & 0.00\(_{0.00}\)  & 67.45  & 7 \\
    TVAE  & 96.71\(_{1.67}\)  & 84.11\(_{1.76}\)  & 62.29\(_{7.40}\)  & 82.90\(_{0.92}\)  & 90.58\(_{5.51}\)  & 89.40\(_{8.92}\)  & 84.33  & 4 \\
    TabDDPM & 96.07\(_{0.34}\)  & 97.73\(_{0.43}\)  & 92.53\(_{1.74}\)  & 98.20\(_{0.73}\)  & \textbf{98.67\(_{0.94}\)}  & 0.00\(_{0.00}\)  & 69.66  & 6 \\
    TABSYN & 94.34\(_{1.36}\)  & \textbf{98.96\(_{0.10}\)}  & 95.09\(_{0.40}\)  & 90.93\(_{0.19}\)  & 91.11\(_{0.41}\)  & \textbf{99.30\(_{0.05}\)}  & 94.95  & 3 \\
    TABDIFF & \underline{99.41\(_{0.34}\)}  & 98.38\(_{0.47}\)  & \underline{99.28\(_{0.49}\)}  & \textbf{99.32\(_{0.34}\)}  & \underline{98.59\(_{0.10}\)}  & 92.83\(_{2.25}\)  & \underline{97.97} & 2 \\
    \midrule
    \ours\  & \textbf{99.55\(_{0.14}\)} & \underline{98.62\(_{0.20}\)}  & \textbf{99.41\(_{0.13}\)}  & \underline{99.02\(_{0.18}\)}  & 96.40\(_{0.24}\)  & \underline{98.38\(_{1.11}\)}  & \textbf{98.56}  & 1 \\
    \bottomrule
    \end{tabular}%
    }
  \label{tab:alp}%
\end{table*}%

\begin{table*}[htbp]
  \centering
  \caption{Comparison of \(\beta\)-Recall scores. Bold Face represents the best score on each dataset and marks the suboptimal ones with an underline.}
  \resizebox{0.95\textwidth}{!}{
    \begin{tabular}{ccccccccc}
    \toprule
    Methods & Adult & Default & Shoppers & Magic & Beijing & News  & Average & Ranking \\
    \midrule
    CTGAN & 30.24\(_{1.19}\)  & 19.77\(_{0.64}\)  & 32.77\(_{0.84}\)  & 10.67\(_{1.35}\)  & 40.17\(_{1.42}\)  & 26.18\(_{1.02}\)  & 26.63  & 6 \\
    CTGAN+ & 29.64\(_{7.37}\)  & 21.59\(_{11.59}\)  & 17.23\(_{3.02}\)  & 0.17\(_{0.02}\)  & 41.65\(_{1.99}\)  & 0.00\(_{0.00}\)  & 18.38  & 7 \\
    TVAE  & 36.67\(_{1.08}\)  & 21.49\(_{0.98}\)  & 23.20\(_{2.65}\)  & 32.47\(_{0.44}\)  & 26.26\(_{2.29}\)  & 27.78\(_{1.17}\)  & 27.98  & 5 \\
    TabDDPM & 48.72\(_{0.75}\)  & \underline{47.16\(_{0.82}\)}  & \textbf{54.13\(_{0.74}\)} & \textbf{47.57\(_{0.93}\)}  & 19.13\(_{0.87}\)  & 0.00\(_{0.00}\)  & 36.12  & 3 \\
    TABSYN & 33.24\(_{0.33}\)  & 39.54\(_{0.26}\)  & 38.66\(_{0.94}\)  & 17.22\(_{0.08}\)  & 20.34\(_{0.15}\)  & \underline{44.47\(_{0.04}\)}  & 32.25  & 4 \\
    TABDIFF & \textbf{49.37\(_{1.21}\)}  & \textbf{50.53\(_{0.82}\)}  & 49.51\(_{1.56}\)  & \underline{46.85\(_{0.79}\)}  & \underline{58.06\(_{0.13}\)}  & 37.32\(_{45.34}\)  & \underline{48.67}  & 2 \\
    \midrule
    \ours\  & \underline{48.90\(_{0.84}\)}  & 45.83\(_{0.22}\)  & \underline{53.09\(_{0.50}\)}  & 46.53\(_{0.57}\)  & \textbf{60.50\(_{0.43}\)}  & \textbf{45.34\(_{0.28}\)}  & \textbf{50.03}  & 1 \\
    \bottomrule
    \end{tabular}%
    }
  \label{tab:beta}%
\end{table*}%
We use column-related distribution, including the column-wise density estimation and pairwise column correlation estimation, to evaluate the fidelity of synthetic data generated from different models. However, these results are insufficient to evaluate the synthetic data’s overall density estimation performance. Therefore, in this section, we adopt \(\alpha\)-Precision and \(\beta\)-Recall to evaluate the joint distribution of the synthetic data.

As shown in Tables~\ref{tab:alp} and~\ref{tab:beta}, we compared the \(\alpha\)-Precision and \(\beta\)-Recall scores of the \ours\ and baseline, respectively. \ours\ achieved improvements of 0.60\% and 2.72\% in these two metrics compared to the suboptimal method TABDIFF~\citep{shi2025tabdiff}, indicating that \ours\ maintains a balance between extensive data coverage and preserving fine-grained details, thus faithfully capturing the breadth and depth of the true data distribution.

\subsection{Detection Score (C2ST)}
\label{suba:c2st}
\begin{table*}[htbp]
  \centering
  \caption{Detection score (C2ST) using logistic regression classifier. Bold Face represents the best score on each dataset and marks the suboptimal ones with an underline.}
    \begin{tabular}{ccccccc}
    \toprule
    Methods & Adult & Default & Shoppers & Magic & Beijing & News \\
    \midrule
    CTGAN & 0.644\(_{0.054}\)  & 0.639\(_{0.025}\)  & 0.711\(_{0.014}\)  & 0.681\(_{0.014}\)  & 0.774\(_{0.078}\)  & 0.695\(_{0.059}\)  \\
    CTGAN+ & 0.436\(_{0.284}\)  & 0.660\(_{0.095}\)  & 0.002\(_{0.001}\)  & 0.020\(_{0.020}\)  & 0.697\(_{0.220}\)  & 0.000\(_{0.000}\)  \\
    TVAE  & 0.703\(_{0.057}\)  & 0.545\(_{0.034}\)  & 0.297\(_{0.020}\)  & 0.821\(_{0.031}\)  & 0.694\(_{0.058}\)  & 0.441\(_{0.023}\)  \\
    TabDDPM & 0.960\(_{0.009}\)  & \textbf{0.981\(_{0.011}\)}  & 0.861\(_{0.006}\)  & \textbf{0.984\(_{0.012}\)}  & 0.350\(_{0.534}\)  & 0.162\(_{0.227}\)  \\
    TABSYN & 0.572\(_{0.019}\)  & 0.715\(_{0.004}\)  & 0.680\(_{0.007}\)  & 0.720\(_{0.003}\)  & 0.671\(_{0.001}\)  & 0.839\(_{0.006}\)  \\
    TABDIFF & \underline{0.967\(_{0.012}\)}  & \underline{0.949\(_{0.008}\)}  & \textbf{0.959\(_{0.029}\)}  & 0.970\(_{0.007}\)  & \textbf{0.960\(_{0.007}\)}  & \underline{0.925\(_{0.029}\)}  \\
    \midrule
    \ours\  & \textbf{0.969\(_{0.012}\)}  & \underline{0.952\(_{0.016}\)}  & \underline{0.946\(_{0.026}\)}  & \underline{0.979\(_{0.024}\)}  & \underline{0.951\(_{0.012}\)}  & \textbf{0.949\(_{0.014}\)}  \\
    \bottomrule
    \end{tabular}%
  \label{tab:c2st}%
\end{table*}%
\ours\ achieves either the best or second-best performance in terms of the Detection Score measured by the C2ST. This metric reflects how distinguishable the synthetic data is from the real data. The strong performance of \ours\ under this metric suggests that the synthetic samples generated are highly realistic and closely align with the underlying real data distribution. Compared to baseline models, \ours\ produces synthetic data that is more difficult for a classifier (logistic regression) to distinguish from real data, demonstrating its effectiveness in preserving key statistical properties while minimizing artifacts that often arise in generative processes. This property is particularly important for downstream applications where distributional consistency between synthetic and real data is crucial.

\subsection{Analysis of Condition Guidance Level Parameters}
\label{sub:para}
\begin{figure*}[t]
\centering
\includegraphics[width=1.0\textwidth]{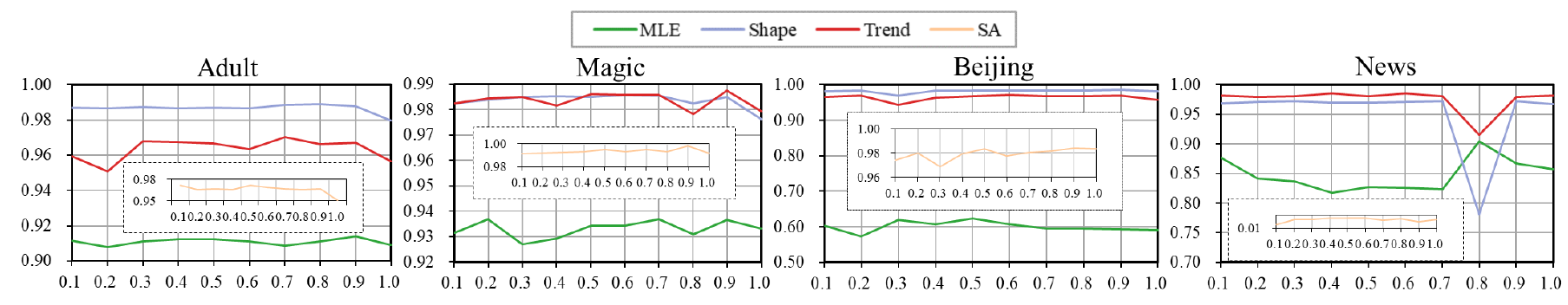}
\caption{\label{figure:abl} The impact of Condition Guidance Level Parameters $p_u$ on MLE, SA, Shape, and Trend across the Adult, Magic, Beijing, and News datasets.
}
\end{figure*}
To verify the impact of different Condition Guidance Levels $p_u$ on the synthesized data (i.e., what extent condition information should be considered during the sampling process), we evaluate the variation trends of MLE, SA, Shape, and Trend with respect to the $p_u$ on four datasets: Adult, Magic, Beijing, and News. Adult and Magic are classification datasets, while Beijing and News are regression datasets.

As shown in Fig.~\ref{figure:abl}, we observe that the MLE, SA, Shape, and Trend metrics exhibit similar trends in response to changes in the Condition Guidance Level Parameter $p_u$. The almost consistent pattern of changes in these metrics suggests that there may be a correlation between them, and that adjusting $p_u$ influences multiple aspects of the synthesized data in a related manner.

\subsection{Analysis of Prior Masking Probability}
\label{app:mask_probability}

We further analyze the effect of the prior masking probability $p_u$. In the main experiments, we set $p_u=0.2$ for all datasets. This setting reflects a practical scenario where tabular schema and textual descriptions may be partially incomplete, while most semantic information remains available. In real-world tabular datasets, semantic priors are unlikely to be almost entirely missing. Therefore, extremely large masking probabilities may correspond to overly pessimistic settings.

The parameter $p_u$ controls the probability of masking prior tokens during training. A larger $p_u$ weakens the semantic guidance received by the denoising network and may reduce semantic consistency. However, the effect is not necessarily strictly monotonic, since moderate masking can also regularize the model and prevent over-reliance on complete priors. Table~\ref{tab:mask_probability_analysis} reports the results under different values of $p_u$. Overall, increasing $p_u$ tends to reduce Shape, Trend, and SA, which is consistent with the intuition that masking more prior tokens weakens semantic guidance during denoising. Nevertheless, the degradation is generally moderate when $p_u \leq 0.2$, supporting the choice of $p_u=0.2$ as a practical default setting.

\begin{table*}[t]
\centering
\caption{Effect of prior masking probability $p_u$ on data fidelity and semantic consistency. The main experiments use $p_u=0.2$.}
\label{tab:mask_probability_analysis}
\setlength{\tabcolsep}{4pt}
\renewcommand{\arraystretch}{1.05}

\begin{minipage}{0.48\textwidth}
\centering
\textbf{Beijing}
\vspace{2pt}

\resizebox{\textwidth}{!}{
\begin{tabular}{lccccc}
\toprule
Metric & $p_u=0$ & $p_u=0.1$ & $p_u=0.2$ & $p_u=0.5$ & $p_u=0.9$ \\
\midrule
Shape(\%) $\uparrow$ & 99.12 & 98.82 & 98.56 & 98.04 & 97.31 \\
Trend(\%) $\uparrow$ & 97.43 & 97.46 & 97.49 & 96.72 & 95.84 \\
SA(\%) $\uparrow$    & 98.20 & 98.75 & 98.68 & 98.21 & 97.36 \\
\bottomrule
\end{tabular}
}
\end{minipage}
\hfill
\begin{minipage}{0.48\textwidth}
\centering
\textbf{News}
\vspace{2pt}

\resizebox{\textwidth}{!}{
\begin{tabular}{lccccc}
\toprule
Metric & $p_u=0$ & $p_u=0.1$ & $p_u=0.2$ & $p_u=0.5$ & $p_u=0.9$ \\
\midrule
Shape(\%) $\uparrow$ & 97.08 & 96.81 & 96.90 & 96.34 & 95.61 \\
Trend(\%) $\uparrow$ & 98.02 & 98.78 & 98.60 & 98.11 & 97.36 \\
SA(\%) $\uparrow$    & 1.71  & 1.46  & 1.24  & 1.08  & 0.82 \\
\bottomrule
\end{tabular}
}
\end{minipage}

\vspace{8pt}

\begin{minipage}{0.48\textwidth}
\centering
\textbf{Shoppers}
\vspace{2pt}

\resizebox{\textwidth}{!}{
\begin{tabular}{lccccc}
\toprule
Metric & $p_u=0$ & $p_u=0.1$ & $p_u=0.2$ & $p_u=0.5$ & $p_u=0.9$ \\
\midrule
Shape(\%) $\uparrow$ & 98.44 & 98.21 & 97.89 & 97.36 & 96.58 \\
Trend(\%) $\uparrow$ & 98.61 & 98.34 & 98.00 & 97.55 & 96.83 \\
SA(\%) $\uparrow$    & 100.00 & 99.59 & 99.98 & 99.74 & 99.21 \\
\bottomrule
\end{tabular}
}
\end{minipage}
\hfill
\begin{minipage}{0.48\textwidth}
\centering
\textbf{Adult}
\vspace{2pt}

\resizebox{\textwidth}{!}{
\begin{tabular}{lccccc}
\toprule
Metric & $p_u=0$ & $p_u=0.1$ & $p_u=0.2$ & $p_u=0.5$ & $p_u=0.9$ \\
\midrule
Shape(\%) $\uparrow$ & 99.52 & 99.31 & 99.06 & 98.63 & 97.91 \\
Trend(\%) $\uparrow$ & 98.55 & 98.22 & 97.98 & 97.36 & 96.58 \\
SA(\%) $\uparrow$    & 94.61 & 94.18 & 93.76 & 92.94 & 91.70 \\
\bottomrule
\end{tabular}
}
\end{minipage}

\vspace{8pt}

\begin{minipage}{0.48\textwidth}
\centering
\textbf{Default}
\vspace{2pt}

\resizebox{\textwidth}{!}{
\begin{tabular}{lccccc}
\toprule
Metric & $p_u=0$ & $p_u=0.1$ & $p_u=0.2$ & $p_u=0.5$ & $p_u=0.9$ \\
\midrule
Shape(\%) $\uparrow$ & 98.68 & 98.82 & 98.57 & 98.01 & 97.24 \\
Trend(\%) $\uparrow$ & 97.67 & 98.12 & 97.89 & 97.35 & 96.49 \\
SA(\%) $\uparrow$    & 99.89 & 100.00 & 100.00 & 99.84 & 99.36 \\
\bottomrule
\end{tabular}
}
\end{minipage}
\hfill
\begin{minipage}{0.48\textwidth}
\centering
\textbf{Magic}
\vspace{2pt}

\resizebox{\textwidth}{!}{
\begin{tabular}{lccccc}
\toprule
Metric & $p_u=0$ & $p_u=0.1$ & $p_u=0.2$ & $p_u=0.5$ & $p_u=0.9$ \\
\midrule
Shape(\%) $\uparrow$ & 99.22 & 98.98 & 98.76 & 98.23 & 97.51 \\
Trend(\%) $\uparrow$ & 99.18 & 98.94 & 98.73 & 98.21 & 97.38 \\
SA(\%) $\uparrow$    & 99.96 & 99.86 & 99.74 & 99.42 & 98.77 \\
\bottomrule
\end{tabular}
}
\end{minipage}

\end{table*}

\subsection{Visualization of Authenticity Comparison between Synthetic Data and Real Data}
\label{suba:auth}

\begin{figure*}[t]
\centering
\includegraphics[width=1.0\textwidth]{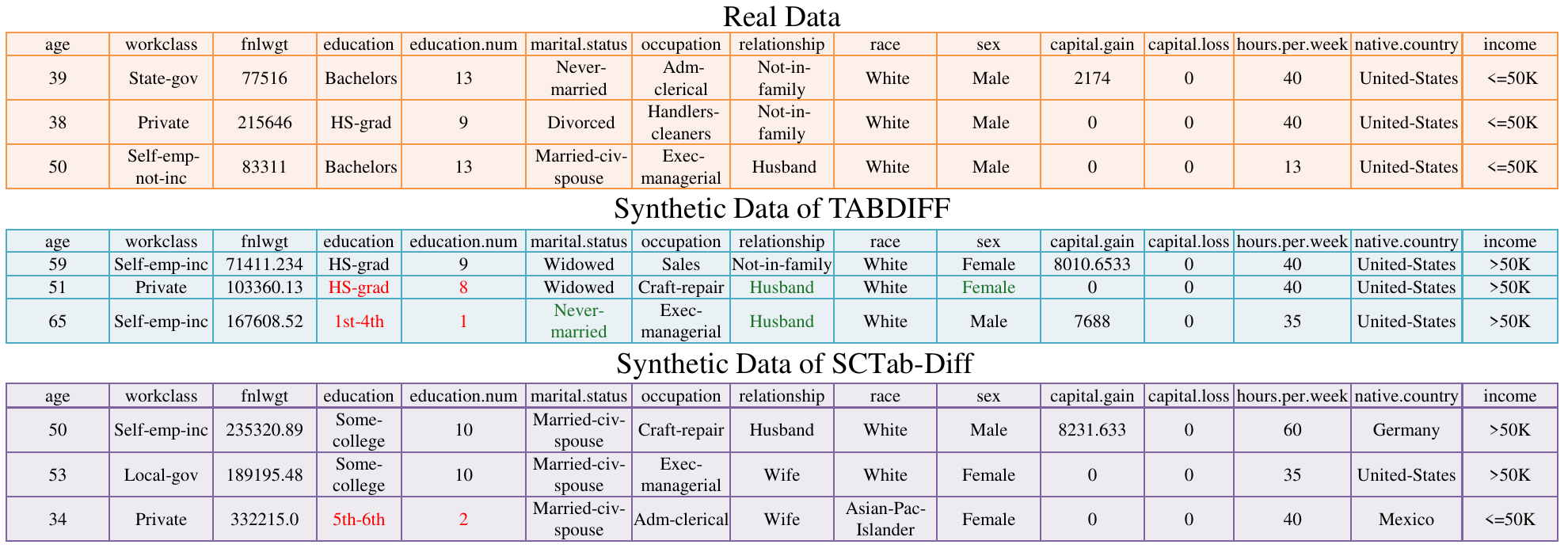}
\caption{\label{figure:vis_exm} Visualization of real and synthetic samples on the Adult dataset.}
\end{figure*}

Fig.~\ref{figure:vis_exm} provides a sample-level authenticity comparison between real Adult samples and synthetic samples generated by TABDIFF and \ours. Unlike distribution-level metrics, this visualization directly examines whether individual generated rows satisfy weak semantic priors. In the figure, highlighted cells within the same row indicate detected violations of intra-column semantics or inter-column symbolic rules. These violations correspond to unrealistic local patterns that may not be fully reflected by aggregate distributional fidelity metrics.

As shown in Fig.~\ref{figure:vis_exm}, real samples are generally consistent with the predefined weak semantic priors. For example, categorical attributes take valid values, numerical attributes remain within plausible ranges, and related columns such as \emph{education} and \emph{education\_num} follow their expected correspondence. In contrast, TABDIFF can generate rows that appear plausible at the distribution level but contain tuple-level semantic inconsistencies. Typical examples include mismatches between \emph{education} and \emph{education\_num}, as well as incompatible combinations among \emph{marital.status}, \emph{relationship}, and other demographic attributes. These errors indicate that relying only on implicit correlations learned from data may be insufficient for enforcing semantic validity at the row level.

Compared with TABDIFF, \ours\ produces fewer highlighted violations and better preserves the logical consistency among related columns. This improvement comes from the prior-conditioned denoising process, where weak semantic priors are encoded in the unified semantic space and injected into generation as semantic conditions. As a result, \ours\ is better able to reduce unrealistic combinations while still maintaining distributional plausibility. This sample-level visualization further supports the quantitative SA results and shows that \ours\ improves not only statistical realism, but also semantic consistency of generated tabular rows.

\section{Detailed Visualization Experiment}
\label{suba:vis}
This section supplements Section~\ref{sub:cat_vis} by providing a more detailed description of the distribution visualization experiment, as outlined below:


\subsection{Detailed Description of Numerical Column Distribution Visualization}
\label{suba:vis_exm2}

\begin{figure*}[t]
\centering
\includegraphics[width=1.0\textwidth]{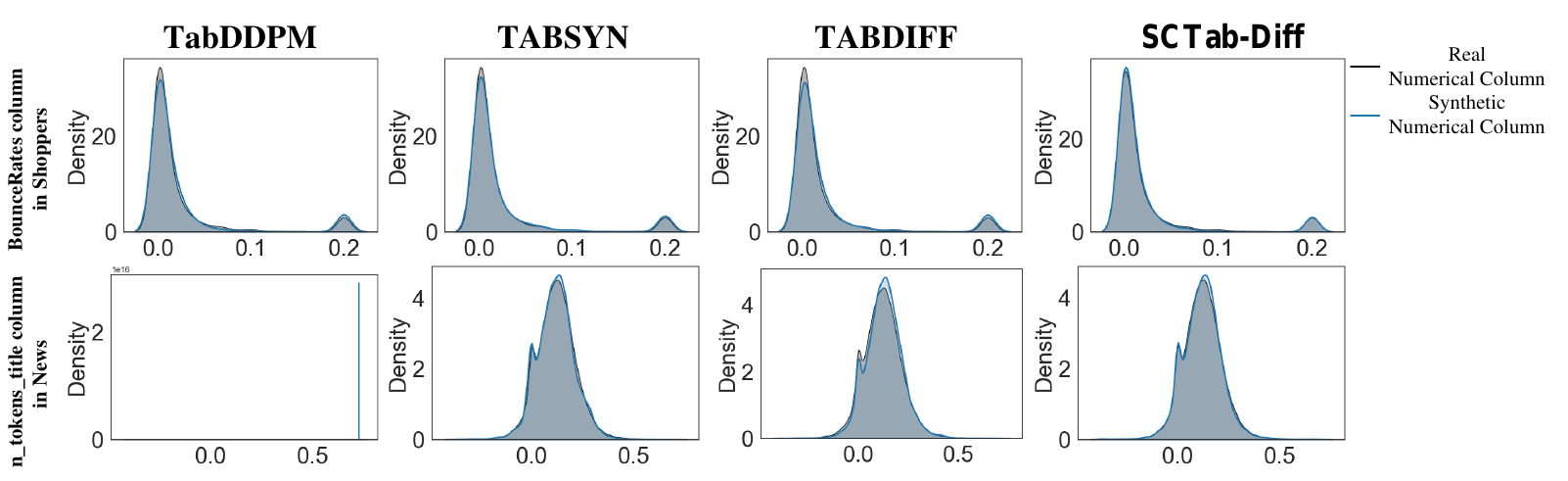}
\caption{\label{figure:vis_exm2} Visualization of intra-column distribution fidelity on Shoppers and News.}
\end{figure*}

Fig.~\ref{figure:vis_exm2} provides a fine-grained visualization of numerical column distribution fidelity. Specifically, we plot the one-dimensional kernel density estimation (KDE) curves for the numerical column \emph{BounceRates} in Shoppers and the numerical column \emph{n\_tokens\_title} in News. The black curve denotes the real data distribution, while the blue curve denotes the synthetic data distribution generated by each method. This visualization complements the Shape metric by showing whether each model can preserve detailed numerical density patterns beyond aggregate column-wise scores.

As shown in Fig.~\ref{figure:vis_exm2}, different methods exhibit clear differences in numerical distribution preservation. For the \emph{BounceRates} column in Shoppers, the real distribution is highly concentrated near zero with a small peak around larger values. TabDDPM, TABSYN, and TABDIFF fail to accurately capture these density peaks, which indicates that they either over-smooth the numerical distribution or shift probability mass away from high-density regions. In contrast, \ours\ more closely follows the real KDE curve and better preserves both the dominant near-zero region and the secondary high-value region.

For the \emph{n\_tokens\_title} column in News, the distribution is more challenging because News contains a relatively large number of features. TabDDPM exhibits a severe degeneration on this column, where the generated values are highly concentrated, and most synthetic samples collapse into a narrow range. As a result, its KDE curve becomes extremely sharp, making the real distribution visually compressed under the same density scale. This suggests that TabDDPM struggles to generate high-dimensional tabular data with many correlated features, possibly because it does not explicitly model feature identity and cross-feature dependency during denoising.
In contrast, TABSYN and TABDIFF produce more reasonable distributions, while \ours\ further aligns better with the real density shape. This indicates that the unified semantic space and prior-conditioned denoising help \ours\ preserve numerical column distributions even when the table contains many features.

For the \emph{n\_tokens\_title} column in News, the distribution is more complex and has a clear central density region. TabDDPM exhibits a visible distributional shift and fails to match the real density shape. TABSYN and TABDIFF improve the fit, but still show mismatches around the main peak. \ours\ produces a synthetic distribution that is more aligned with the real curve. The unified semantic space and column-wise diffusion process help preserve value-sensitive numerical semantics. 

\subsection{Detailed Description of Inter-Column Dependency Visualization}
\label{suba:vis_exm3}

\begin{figure*}[t]
\centering
\includegraphics[width=0.8\textwidth]{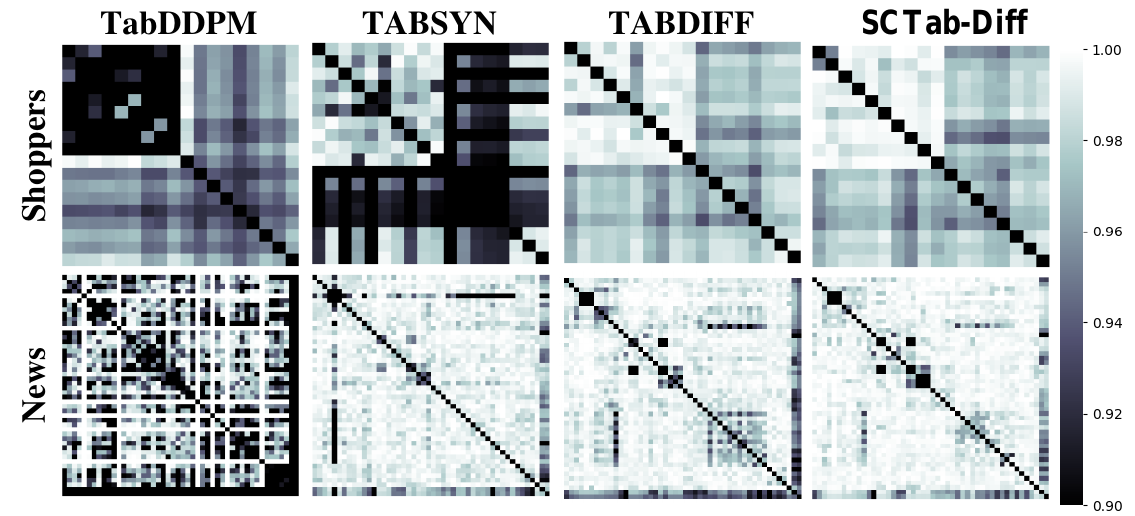}
\caption{\label{figure:vis_exm3} Visualization of inter-column dependency preservation on Shoppers and News.}
\vspace{-10pt}
\end{figure*}

Fig.~\ref{figure:vis_exm3} visualizes the ability of different methods to preserve inter-column dependency patterns on Shoppers and News. We use heatmaps to compare pair-wise dependency preservation between real and synthetic data. Lighter colors indicate stronger preservation of real dependency patterns, while darker regions indicate larger deviations. This visualization complements the Trend metric by showing whether a model can maintain global cross-column structures rather than only individual column distributions.

As shown in Fig.~\ref{figure:vis_exm3}, TabDDPM and TABSYN show more dark blocks on both datasets. This suggests that their generated data contains larger discrepancies in pair-wise column relationships. These deviations are especially visible on News, where the number of columns is larger and the dependency structure is more complex. Although TABDIFF preserves part of the dependency structure, it still shows visible mismatches in several off-diagonal regions.

In contrast, \ours\ produces heatmaps that are more consistent with the real dependency patterns. On Shoppers, \ours\ better preserves the block-wise dependency structure among behavioral features. On News, \ours\ maintains more coherent pair-wise relationships across article-level features. This improvement can be attributed to prior-conditioned denoising, where weak semantic priors and cross-column context are jointly used during the reverse process. These results indicate that \ours\ not only preserves marginal column distributions, but also better captures inter-column dependencies. This further supports the Trend improvements reported in Table~\ref{tab:data_fidelity}.

\section{Limitations}
\label{sec:limitations}

\ours\ has several limitations. First, the quality of weak semantic priors depends on the availability and informativeness of tabular schema and textual descriptions. Although candidate priors are validated on the real training split, noisy or incomplete metadata may still lead to imperfect semantic guidance. Second, the current framework mainly considers intra-column semantics and inter-column symbolic rules. More complex forms of domain knowledge, such as causal relations or procedural constraints, are not explicitly modeled. Third, our experiments are conducted on six real-world tabular benchmarks, which may not fully cover highly specialized domains with stricter semantic requirements. Finally, while the LLM used for prior extraction does not access raw tabular records, the generative model is still trained on real data. Future work will explore stronger privacy auditing, privacy-preserving training, and richer semantic prior representations.

\tcbset{
    colback=white!10,
    colframe=black,
    boxrule=0.3mm,
    arc=2mm,
    fonttitle=\bfseries,
    left=2mm, right=2mm,
    top=1mm, bottom=1mm  
}
\begin{table*}[h]
 \centering
  \caption{\label{figure:instruction}Weak Semantic Priors of Adult.}
\begin{tcolorbox}
    Dataset Constraints \\
    \textcolor[rgb]{ .929,  .49,  .192}{Predict whether the annual income of an individual exceeds \$50K/yr based on census data. Also known as the ``Census Income'' dataset.} \\
    \\
    Intra-Column Semantics \\
    \textcolor[rgb]{ .357,  .608,  .835}{1.Age is between 17 and 90 (inclusive of 17, exclusive of 91).\\
2.Workclass must be one of a specific list (e.g., ``Private'', ``Federal-gov'', etc.).\\
3.Fnlwgt (final weight) must be a non-negative number.\\
4.Education must be from a predefined list (e.g., ``HS-grad'', ``Masters'', etc.).\\
5.Education number (numeric representation) must be between 1 and 16.\\
6.Marital status must match one of the listed valid options.\\
7.Occupation must match a set list of roles.\\
8.Relationship must be a valid type (e.g., ``Husband'', ``Own-child'').\\
9.Race, Sex, and Income must be one of the specified categories.\\
10.Capital gain must be between 0 and 99,999.\\
11.Capital loss must be non-negative.\\
12.Hours worked per week must be between 1 and 99.\\
13.Native country must be one of the recognized entries.} \\ \\
    Inter-Column Symbolic Rules \\
    \textcolor[rgb]{ .494,  .392,  .62}{1.If the relationship is ``Husband'' or ``Wife'', then marital status must imply the person is or was married. If someone is ``Never-married'', they shouldn't be listed as ``Husband'' or ``Wife''.\\
2.if someone has a higher education degree (Masters, Doctorate, Prof-school), they must be at least 22 years old.\\
3.Ensures that if the workclass indicates self-employment, the relationship should not be ``Own-child''.}
\end{tcolorbox}

\label{tab:adult}%
\end{table*}%

\begin{table*}[h]
 \centering
  \caption{\label{figure:instruction}Weak Semantic Priors of Shoppers.}
\begin{tcolorbox}
    Dataset Constraints \\
    \textcolor[rgb]{ .929,  .49,  .192}{Of the 12,330 sessions in the dataset, 84.5\% (10,422) were negative class samples that did not end with shopping, and the rest (1908) were positive class samples ending with shopping.} \\
    \\
    Intra-Column Semantics \\
    \textcolor[rgb]{ .357,  .608,  .835}
{1.Administrative, Informational, ProductRelated: must be non-negative integers within known upper limits (e.g.,Administrative \(\leq\) 27, Informational \(\leq\) 24, ProductRelated \(\leq\) 705).\\
2.Durations: (e.g., Administrative\_Duration) must be non-negative.\\
3.BounceRates and ExitRates: must be between 0 and 0.2.\\
4.PageValues and SpecialDay: must be \(\geq\) 0, with SpecialDay \(\leq\) 1.\\
5.Month: must be a valid month string (e.g., “May”, “Nov”).} \\ \\
    Inter-Column Symbolic Rules \\
    \textcolor[rgb]{ .494,  .392,  .62}{1.If a visit count (e.g., Administrative) is zero, the corresponding duration must also be zero.\\
2.If duration is greater than 0, then the corresponding visit count must be \(>\) 0.\\
3.If BounceRates \(>\) 0.2, then ProductRelated should be \(\leq\) 10.\\
4.If a user viewed fewer than one product-related page, then PageValues must be 0.\\
5.If PageValues \(>\) 0, then ProductRelated\_Duration \(>\) 0 must be true.}
\end{tcolorbox}

\label{tab:addlabel}%
\end{table*}%
 
\begin{table*}[h]
 \centering
  \caption{\label{figure:instruction}Weak Semantic Priors of Default.}
\begin{tcolorbox}
    Dataset Constraints \\
    \textcolor[rgb]{ .929,  .49,  .192}{The dataset consists of feature vectors belonging to 12,330 sessions. The dataset was formed so that each session would belong to a different user in a 1-year period to avoid any tendency to a specific campaign, special day, user profile, or period.} \\
    \\
    Intra-Column Semantics \\
    \textcolor[rgb]{ .357,  .608,  .835}{
1.LIMIT\_BAL: Credit limit is between 10,000 and 1,000,000.\\
2.SEX: Coded as 1 (male) or 2 (female).\\
3.EDUCATION: Must be one of 0–6 (0 = unknown, 1 = graduate school, etc.).\\
4.MARRIAGE: Must be one of 0–3 (0 = unknown, 1 = married, etc.).\\
5.AGE: Between 18 and 100 years.\\
6.PAY\_0 to PAY\_6: Monthly payment statuses must be in the range [-2, 8]. These indicate repayment behavior (e.g., -1 = paid in full, 1 = one month delay, etc.).\\
7.BILL\_AMT1 to BILL\_AMT6: Monthly bill amounts fall within observed realistic ranges (can be negative, possibly indicating credit).\\
8.PAY\_AMT1 to PAY\_AMT6: Payments are between 0 and an upper bound (no negative payments).\\
9.default\_payment\_next\_month: Must be 0 or 1 (0 = no default, 1 = default).} \\ \\
    Inter-Column Symbolic Rules \\
    \textcolor[rgb]{ .494,  .392,  .62}{1.For any month where the payment status (PAY\_*) indicates a delay greater than 1 month, the corresponding bill amount (BILL\_AMT*) must be non-negative (i.e., the person owed money)\\
2.If there is any positive payment in a month, the total of all bill amounts should be greater than or equal to the total payments.\\
3.If a person is 20 years or younger, their credit limit (LIMIT\_BAL) must be zero or less (suggesting they should not have a credit limit).}
\end{tcolorbox}

\label{tab:addlabel}%
\end{table*}%

\begin{table*}[h]
 \centering
  \caption{\label{figure:instruction}Weak Semantic Priors of Magic.}
\begin{tcolorbox}
    Dataset Constraints \\
    \textcolor[rgb]{ .929,  .49,  .192}{Data are MC generated to simulate registration of high-energy gamma particles in an atmospheric Cherenkov telescope.} \\
    \\
    Intra-Column Semantics \\
    \textcolor[rgb]{ .357,  .608,  .835}
{1.All physical quantities (like Length, Width, Size, Dist) must be non-negative.\\
2.Concentration ratios (Conc, Conc1) must be between 0 and 1.\\
3.Asymmetry and moment features must fall within realistic range limits (e.g., -500 to 580).\\
4.Alpha (angle in degrees) must be between 0 and 90.\\
5.class must be either 'g' (gamma-ray signal) or 'h' (hadron background).
} \\ \\
    Inter-Column Symbolic Rules \\
    \textcolor[rgb]{ .494,  .392,  .62}{
1.The major axis (Length) must be \(\geq\) minor axis (Width), as expected for an ellipse.\\
2.The intensity from the highest pixel (Conc1) must be \(\leq\) the sum of the top 2 pixels (Conc), and both must be \(\leq\)1.\\
3.If the image is highly concentrated (Conc \(>\) 0.5), the total intensity (Size) should be sufficiently large (\(>\)1.5). This helps avoid extreme concentration in very faint showers, which is likely noise.\\
4.If the ellipse is centered (Dist \(<\) 50) and symmetric (Asym \(\approx\) 0), then the orientation angle (Alpha) should be small (\(<\) 30°). This reflects physically aligned gamma-ray events.\\
5.If the asymmetry is large (absolute value \(>\) 100 mm), then the third moment along the major axis (M3Long) should also be large (\(>\)\(|8|\) mm), reflecting skewed energy distributions in real air showers.}
\end{tcolorbox}

\label{tab:addlabel}%
\end{table*}%

\begin{table*}[h]
 \centering
  \caption{\label{figure:instruction}Weak Semantic Priors of Beijing.}
\begin{tcolorbox}
    Dataset Constraints \\
    \textcolor[rgb]{ .929,  .49,  .192}{This data set contains the PM2.5 data of the US Embassy in Beijing. Meanwhile, meteorological data from Beijing Capital International Airport are also included.} \\
    \\
    Intra-Column Semantics \\
    \textcolor[rgb]{ .357,  .608,  .835}
{1.Year should be between 2010 and 2014.\\
2.Month: between 1 and 12, Day: between 1 and 31, Hour: between 0 and 23.\\
3.PM2.5 concentration (pm2\_5) must be non-negative.\\
4.Dew point (DEWP) must be \(\geq\) -100.\\
5.Temperature must be between -50 and 60 °C.\\
6.Pressure (PRES) must be between 800 and 1100 hPa.\\
7.Wind direction (cbwd) must be one of ('SE', 'NW', 'cv', 'NE').\\
8.Wind speed (Iws) and other count columns (Is, Ir) must be non-negative.
} \\ \\
    Inter-Column Symbolic Rules \\
    \textcolor[rgb]{ .494,  .392,  .62}{1.Ensures that a month like 2 4 6 9 11 doesn't have 31 days.\\
2.For February, it checks if the year is a leap year, and sets the maximum allowed day to 29 (leap) or 28 (non-leap).\\
3.The dew point (DEWP) should never exceed the actual temperature (TEMP).\\
4.If wind speed (Iws) is high (\(>\)10), the PM2.5 (pm2\_5) level is expected to be below 800 (because wind typically disperses pollutants).\\
5.If there is no wind (wind speed (Iws) = 0), then the wind direction (cbwd) should be 'cv' (likely short for “calm/variable”).}
\end{tcolorbox}

\label{tab:addlabel}%
\end{table*}%

\begin{table*}[h]
 \centering
  \caption{\label{figure:instruction}Weak Semantic Priors of News.}
\begin{tcolorbox}
    Dataset Constraints \\
    \textcolor[rgb]{ .929,  .49,  .192}{This dataset summarizes a heterogeneous set of features about articles published by Mashable in a period of two years. The goal is to predict the number of shares in social networks (popularity).} \\
    \\
    Intra-Column Semantics \\
    \textcolor[rgb]{ .357,  .608,  .835}
{
1.Numeric Features (e.g., n\_tokens\_title, average\_token\_length) must be non-negative.\\
2.num\_keywords: must be between 1 and 10.\\
3.LDA\_00 to LDA\_04: must be between 0 and 1.\\
4.title\_sentiment\_polarity, avg\_negative\_polarity, etc., must lie in valid sentiment ranges (e.g., [-1, 1]).\\
5.Share\-related features (e.g., shares, self\_reference\_min\_shares): must lie between 0 and 843300 (the observed max).
} \\ \\
    Inter-Column Symbolic Rules \\
    \textcolor[rgb]{ .494,  .392,  .62}{1.Ensures that the sum of positive and negative word rates is approximately 1 (±0.01 tolerance).\\
2.when there is at least one keyword, the three keyword minimum/average features are each \(\geq\) -1.\\
3.Enforces that the average of keyword metrics follows a logical order: kw\_min\_avg \(\leq\) kw\_avg\_avg \(\leq\) kw\_max\_avg.\\
4.Self-reference shares follow: min \(\leq\) avg \(\leq\) max.\\
5.Verifies that the sum of all LDA topic probabilities is approximately 1.\\
6.Ensures: min\_positive \(\leq\) avg\_positive \(\leq\) max\_positive \\
7.Ensures: min\_negative \(\leq\) avg\_negative \(\leq\) max\_negative\\
8.Ensures: abs\_title\_sentiment\_polarity == abs(title\_sentiment\_polarity)\\
9.If global\_rate\_positive\_words \(>\) 0, then avg\_positive\_polarity must be \(>\) 0.\\
10.If global\_rate\_negative\_words \(>\) 0, then avg\_negative\_polarity must be \(<\) 0.}
\end{tcolorbox}

\label{tab:news}%
\end{table*}%

\FloatBarrier 

\newpage

\end{document}